\documentclass{article}

\PassOptionsToPackage{numbers, compress}{natbib}

\usepackage[preprint]{neurips_2022}

\usepackage[utf8]{inputenc} 
\usepackage[T1]{fontenc}    
\usepackage{hyperref}       
\usepackage{url}            
\usepackage{booktabs}       
\usepackage{amsfonts}       
\usepackage{nicefrac}       
\usepackage{microtype}      
\usepackage[table]{xcolor}
\usepackage{xcolor}         

\usepackage{graphicx}
\usepackage{amsmath}
\usepackage{amssymb}
\usepackage{array}
\usepackage{multirow}
\usepackage{wrapfig}
\usepackage{amssymb}

\usepackage{algorithm}
\usepackage{algpseudocode}

\usepackage[lofdepth,lotdepth]{subfig}
\usepackage[title]{appendix}

\DeclareMathOperator*{\argmin}{arg\,min}

\newcommand{\RR}{\mathbb{R}}

\newcommand{\cC}{\mathcal{C}}
\newcommand{\cL}{\mathcal{L}}

\newcommand{\vece}{\mathbf{e}}
\newcommand{\vech}{\mathbf{h}}

\newcommand{\vecm}{\mathbf{m}}

\newcommand{\vecp}{\mathbf{p}}
\newcommand{\vecr}{\mathbf{r}}

\newcommand{\vecu}{\mathbf{u}}
\newcommand{\vecv}{\mathbf{v}}

\newcommand{\vecz}{\mathbf{z}}

\newcommand{\bS}{\mathbf{S}}
\newcommand{\bX}{\mathbf{X}}
\newcommand{\bZ}{\mathbf{Z}}

\newcommand{\bPi}{\mathbf{\Pi}}

\newcommand{\Q}{\mathcal{Q}}
\newcommand{\RQ}{\mathcal{R\mkern-1.5mu Q}}
\newcommand{\ARmodel}{Contextual RQ-Transformer }

\newcommand{\masktok}{[\text{MASK}]}

\title{Draft-and-Revise: Effective Image Generation with Contextual RQ-Transformer}

\author{%
  Doyup Lee\thanks{Equal contribution} \\
  POSTECH, Kakao Brain\\
  \texttt{doyup.lee@postech.ac.kr} \\
  \And
  Chiheon Kim\footnotemark[1] \\
  Kakao Brain\\
  \texttt{chiheon.kim@kakaobrain.com} \\
  \And 
  Saehoon Kim\\
  Kakao Brain\\
  \texttt{shkim@kakaobrain.com}\\
  \And
  Minsu Cho\\
  POSTECH\\
  \texttt{mscho@postech.ac.kr}\\
  \And
  Wook-Shin Han\thanks{Corresponding author}\\
  POSTECH\\
  \texttt{wshan@dblab.postech.ac.kr}\\
}

\begin{document}

\maketitle

\begin{abstract}
Although autoregressive models have achieved promising results on image generation, their unidirectional generation process prevents the resultant images from fully reflecting global contexts. To address the issue, we propose an effective image generation framework of \emph{Draft-and-Revise} with \emph{Contextual RQ-transformer} to consider global contexts during the generation process. As a generalized VQ-VAE, RQ-VAE first represents a high-resolution image as a sequence of discrete code stacks. After code stacks in the sequence are randomly masked, Contextual RQ-Transformer is trained to infill the masked code stacks based on the unmasked contexts of the image. Then, Contextual RQ-Transformer uses our two-phase decoding, \emph{Draft-and-Revise}, and generates an image, while exploiting the global contexts of the image during the generation process. Specifically. in the \emph{draft} phase, our model first focuses on generating diverse images despite rather low quality. Then, in the \emph{revise} phase, the model iteratively improves the quality of images, while preserving the global contexts of generated images. In experiments, our method achieves state-of-the-art results on conditional image generation. We also validate that the Draft-and-Revise decoding can achieve high performance by effectively controlling the quality-diversity trade-off in image generation.
\end{abstract}


\section{Introduction} \label{sec:intro}

Learning discrete representations of images enables autoregressive (AR) models to achieve promising results on high-resolution image generation. 
Here, an image is encoded into a feature map, which is represented as a sequence of discrete codes~\cite{VQGAN,VQVAE} or code stacks~\cite{RQVAE}. 
Then, an AR model generates a sequence of codes in the raster scan order and decodes the codes into an image.
Consequently, AR models show high performance and scalability on large-scale datasets~\cite{VQGAN,RQVAE,DALL-E}.

\begin{wrapfigure}{r}{0.45\textwidth}
\centering
\includegraphics[width=0.45\textwidth, height=0.15\textwidth]{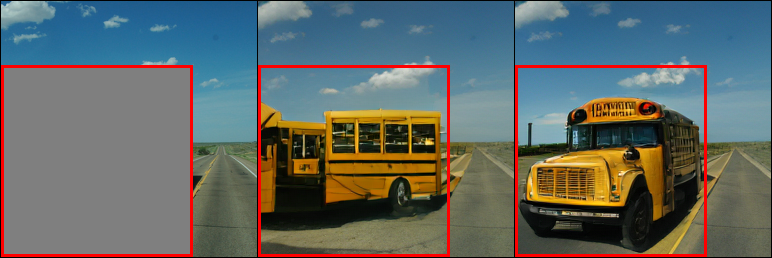}
\caption{Examples of image inpainting by an AR model (middle) and ours (right).}
\label{fig:inpainting}
\vspace{-0.1in}
\end{wrapfigure}

Despite the promising results of AR models, we postulate that the ability of AR models is limited due to the lack of considering global contexts in the generation process. Specifically, since AR models generate images by sequentially predicting the next code and attending to only precedent codes generated, they neither exploit the later part of the generated image nor consider the global contexts during generation.
For example, Figure~\ref{fig:inpainting} (middle) shows that an AR model fails to generate a coherent image, when it is asked to inpaint the masked region of Figure~\ref{fig:inpainting} (left) with a school bus. 
Such a failure is due to the inability of AR models to refer to the context of traffic lane on the right side of the masked region.

To address this issue, we propose an effective image generation framework, \emph{Draft-and-Revise}, with a contextual transformer to exploit the global contexts of images.
Given a randomly masked image, the contextual transformer is first trained to infill the masks by bidirectional self-attentions similarly to BERT~\cite{BERT}.
To fully leverage the contextual prediction in generation, we propose Draft-and-Revise decoding which has two phases, \emph{draft} and \emph{revise}, imitating the image generation process of a human expert who draws a draft first and iteratively revises the draft to improve its quality.
In the draft phase, the model first infills an empty image to generate a draft image with diverse contents despite the rather low-quality. 
In the revise phase, the visual quality of the draft is iteratively improved, while the global contexts of the draft are preserved and exploited.
Consequently, our \emph{Draft-and-Revise} with contextual transformer effectively generates high-quality images with diverse contents.

We use residual-quantized VAE (RQ-VAE)~\cite{RQVAE} to implement our image generation framework, since RQ-VAE generalizes vector-quantized VAE (VQ-VAE)~\cite{VQVAE} by representing an image as a sequence of code stacks instead of a sequence of codes.
Then, we propose Contextual RQ-Transformer as a contextual transformer for masked code stack modeling of RQ-VAE. 
Specifically, given a sequence of randomly masked code stacks, Contextual RQ-Transformer first uses a bidirectional transformer to capture the global contexts of unmasked code stacks.
Based on the global contexts, the masked code stacks are predicted in parallel, while the codes in each masked code stack are sequentially predicted.
In experiments, our Draft-and-Revise framework with Contextual RQ-Transformer achieves state-of-the-art results on conditional image generation and remarkable improvements on image inpainting. 
In addition, we demonstrate that Draft-and-Revise decoding can effectively control the quality-diversity trade-off in image generation to achieve high performance.

The main contributions of this paper are summarized as follows.
1) We propose an intuitive and powerful framework, \emph{Draft-and-Revise}, for image generation based on a bidirectional transformer.
2) We propose Contextual RQ-Transformer for masked code stack modeling of RQ-VAE and empirically show that the proposed model with Draft-and-Revise decoding achieves state-of-the-art results on class- and text-conditional image generation benchmarks.
3) An extensive ablation study validates the effectiveness of Draft-and-Revise decoding on controlling the quality-diversity trade-off and its capability to generate high-quality images with diverse contents.


\section{Related Work}
\paragraph{Discrete Representation for Image Generation}
By representing an image as a sequence of codes, VQ-VAE~\cite{VQVAE} becomes an important part for high-resolution image generation~\cite{MaskGIT,cogview,VQDiff,RQVAE,DALL-E,VQVAE}, but suffers from low quality of reconstructed images.
However, VQGAN~\cite{VQGAN} significantly improves the perceptual quality of reconstructed images by adding the adversarial and perceptual losses into the training objective of VQ-VAE.
As a generalized approach of VQ-VAE and VQGAN, RQ-VAE~\cite{RQVAE} represents an image as a sequence of code stacks, which consists of ordered codes, and reduces the sequence length, while preserving the reconstruction quality.
Then, RQ-Transformer~\cite{RQVAE} achieves high performance with lower computational costs on generating high-resolution images.
However, as an AR model of RQ-VAE, RQ-Transformer cannot capture the global contexts of generated images.

\paragraph{Generation Tasks with Bidirectional Transformers}
To overcome the limitation of AR models on unidirectional architecture, bidirectional transformers have been used for generative tasks.
Similar to the pretraining objective of BERT~\cite{BERT}, a bidirectional transformer is trained to infill a random mask.
Then, accompanied with an iterative decoding method~\cite{mask-predict,GLAT,SUNDAE,BERT-speak}, the model can generate texts~\cite{mask-predict}, images~\cite{MaskGIT}, or videos~\cite{ShowMeWhat,M6UFC}.
Recently, discrete diffusion models~\cite{D3PM,UnleashingTransformer,ImageBART,VQDiff} also uses bidirectional transformers to generate an image.
Given a partially corrupted by random code replacement~\cite{D3PM,ImageBART} or randomly masked~\cite{D3PM,UnleashingTransformer,VQDiff} sequence of codes, diffusion models are trained to gradually denoise the corrupted codes or infill the masks.
The training of discrete diffusion models with an absorbing state~\cite{D3PM} is the same to infill randomly masked sequence~\cite{MaskGIT,UnleashingTransformer}.
However, different from the reverse process of diffusion models, our decoding method has explicit two phases to generate high-quality images with diverse contents.

\section{\emph{Draft-and-Revise} Framework for Effective Image Generation} \label{sec:methods}
\begin{figure}
\centering
\includegraphics[width=\textwidth]{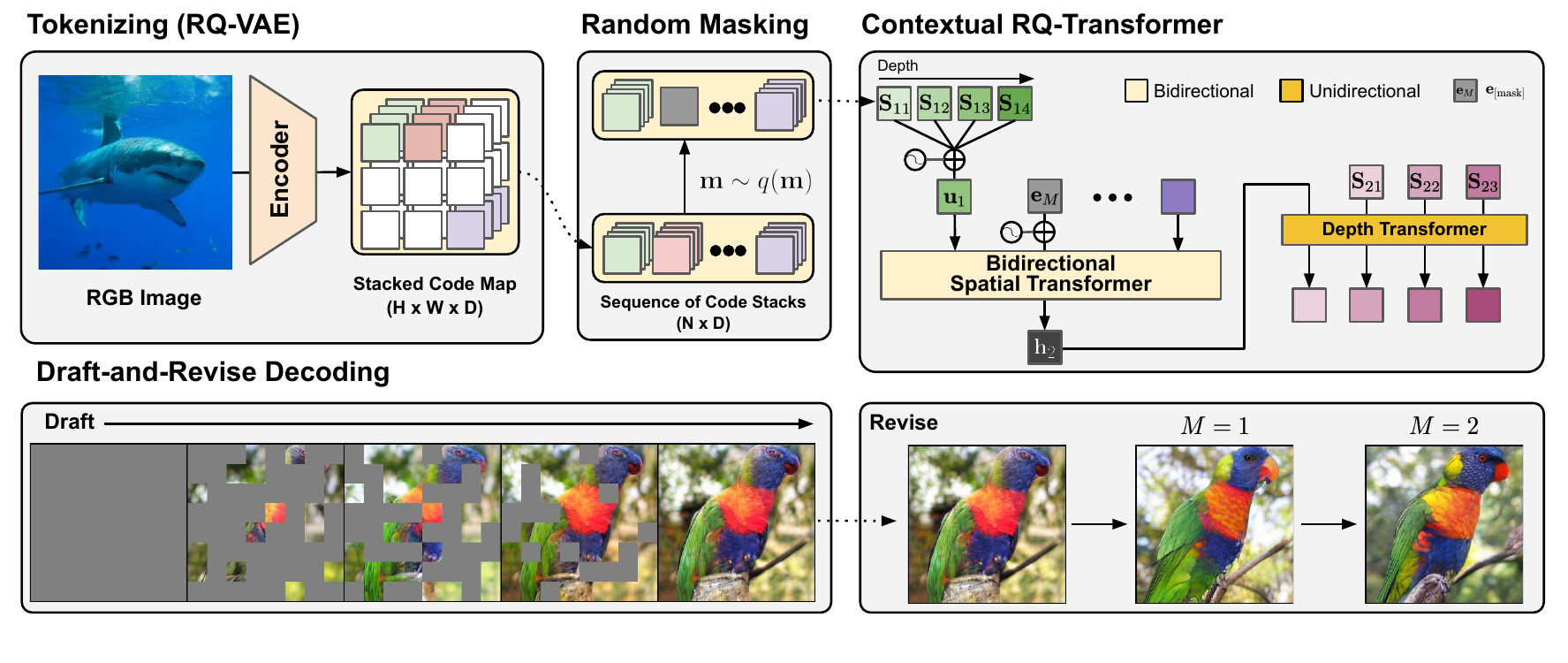}
\caption{The overview of Draft-and-Revise framework with Contextual RQ-Transformer. Our framework exploits global contexts of images to generate high-quality images with diverse contents.}
\label{fig:overview}
\end{figure}

In this section, we propose our \emph{Draft-and-Revise} framework for effective image generation using bidirectional contexts of images. 
We first review RQ-VAE~\cite{RQVAE} as a generalization of VQ-VAE.
Then, we propose Contextual RQ-Transformer which is trained to infill a randomly masked sequence of code stacks of RQ-VAE by understanding bidirectional contexts of unmasked parts in the sequence.
Lastly, we propose \emph{draft-and-revise} decoding for a bidirectional transformer to effectively generate high-quality images exploiting global contexts of images. Figure \ref{fig:overview} provides the overview of our proposed framework, including Contextual RQ-Transformer and Draft-and-Revise decoding.

\subsection{Residual-Quantized Variational Autoencoder (RQ-VAE)}
RQ-VAE~\cite{RQVAE} represents an image as a sequence of code stacks.
Let a codebook $\cC=\{(k, \vece(k))\}_{k \in [K]}$ include pairs of a code $k$ and its code embedding $\vece(k) \in \RR^{n_z}$, where $K=|\cC|$ is the codebook size and $n_z$ is the dimensionality of $\vece(k)$.
Given a vector $\vecz \in \RR^{n_z}$, $\Q(\vecz;\cC)$ is defined as the code of $\vecz$:
\begin{equation}
\label{eq:Q}
    \Q(\vecz; \cC) = \argmin_k \left\|\vecz - \vece(k)\right\|_2^2.
\end{equation}
Then, RQ with depth $D$ represents a vector as a \emph{code stack} which consists of $D$ codes:
\begin{equation}
\label{eq:RQ}
    \RQ (\vecz; \cC, D) = (k_1, \cdots, k_D)\in [K]^D,
\end{equation}
where $k_d$ is the $d$-th code of $\vecz$.
Specifically, RQ first initializes the $0$-th residual vector as $\vecr_0 = \vecz$, and then recursively discretizes a residual vector $\vecr_{d-1}$ and computes the next residual vector $\vecr_d$ as
\begin{equation} \label{eq:RQ_details}
    k_d = \Q(\vecr_{d-1}; \cC),
    \quad\text{}\quad
    \vecr_d = \vecr_{d-1} - \vece(k_d),
\end{equation}
for $d \in [D]$.
Finally, $\vecz$ is approximated by the sum of the $D$ code embeddings $\hat{\vecz}:=\sum_{d=1}^{D}{\vece(k_d)}$.
We remark that RQ is a generalized version of VQ, as RQ with $D\!=\!1$ is equivalent to VQ.
For $D>1$, RQ conducts a finer approximation of $\vecz$ as the quantization errors are sequentially reduced as $d$ increases.
Here, the coarse-to-fine approximation ensures the $D$ codes to be sequentially dependent. 

RQ-VAE represents an image as a map of code stacks.
Specifically, a given image $\bX$ is first converted to a low-resolution feature map $\bZ = E(\bX) \in \RR^{H \times W \times n_z}$, and then each feature vector $\bZ_{hw}$ at spatial position $(h, w)$ is discretized into a code stack by RQ with depth $D$. As a result, we get a map of code stacks $\bS \in [K]^{H \times W \times D}$.
Further details of RQ-VAE are referred to Appendix.


\subsection{Contextual Transformer for Image Generation with Global Contexts}
As a bidirectional transformer for RQ-VAE, we propose Contextual RQ-Transformer for image generation based on a contextual understanding of images.
First, we adopt the pretraining of BERT~\cite{BERT} to formulate a masked code stack modeling of RQ-VAE.
Then, we introduce how Contextual RQ-Transformer infills the randomly masked code stacks after reading the given contextual information.

\subsubsection{Masked Code Stack Modeling of RQ-VAE}
By adopting the pretraining of BERT~\cite{BERT}, we formulate the masked code stack modeling of RQ-VAE with a contextual transformer to generate an image by iterative mask-infilling as non-AR models~\cite{mask-predict}.
We first convert the map $\bS \in [K]^{H \times W \times D}$ into a sequence of code stacks $\bS' \in [K]^{N \times D}$ using the raster-scan ordering, where $N = HW$ and $\bS'_n = (\bS'_{n1}, \cdots, \bS'_{nD}) \in [K]^D$ for $n \in [N]$.
We denote $\bS'$ as $\bS$ for the brevity of notation.
A mask vector $\vecm$ is defined as a binary vector $\vecm \in \{0, 1\}^N$ to indicate the spatial positions to be masked.
Then, the masked sequence $\bS_{\setminus \vecm}$ of $\bS$ by $\vecm$ is defined as
\begin{equation}
    (\bS_{\setminus\vecm})_{n} = 
    \begin{cases}
        \bS_{n} & \text{if $\vecm_n = 0$} \\
        \masktok^D & \text{if $\vecm_n = 1$}
    \end{cases},
\end{equation}
where $\masktok$ is a mask token to substitute for $\bS_{nd}$ if $\vecm_n=1$.
Given a random mask vector $\vecm \sim q(\vecm)$, the masked code stacks given $\bS_{\setminus \vecm}$ are modeled as
\begin{equation} \label{eq:local_AR_mask}
    \prod_{n:\vecm_n=1} p(\bS_{n} | \bS_{\setminus \vecm})
     \\
    = \prod_{n:\vecm_n=1} \prod_{d=1}^{D} p(\bS_{nd} | \bS_{n, <d}, \bS_{\setminus \vecm}),
\end{equation}
where $q(\vecm)$ is a mask distribution where the masking portion $\sum_{n=1}^{N} \vecm_i / N$ in $(0,1]$ as well as the masking positions are randomly chosen.
Instead of fixing the portion to 15\% as in BERT, training a model with a random masking portion from $(0,1]$ enables the model to generate new images based on various masking patterns including $\vecm_n=1$ for all $n$.
We explain the details of $q(\vecm$) in Section~\ref{sec:training_part}.

The left-hand side of Eq.~\ref{eq:local_AR_mask} implies that all masked code stacks can be decoded in parallel, after extracting contextual information from $\bS_{\setminus \vecm}$.
If $D=1$, Eq.~\ref{eq:local_AR_mask} becomes equivalent to conventional masked token modeling of texts~\cite{BERT} and images~\cite{MaskGIT,ShowMeWhat} where a single token at each masked position is predicted.
For $D>1$, the $D$ codes of $\bS_n$ are autoregressively predicted, as they are sequentially computed in Eq.~\ref{eq:RQ_details} for a coarse-to-fine approximation and hence well-suited for an AR prediction.



\subsubsection{Contextual RQ-Transformer}\label{sec:RQ-Transformer}
We modify the previous RQ-Transformer~\cite{RQVAE} for masked code stack modeling with bidirectional contexts in Eq.~\ref{eq:local_AR_mask}.
Contextual RQ-Transformer consists of \emph{Bidirectional Spatial Transformer} and \emph{Depth Transformer}:
Bidirectional Spatial Transformer understands contextual information in the unmasked code stacks using bidirectional self-attentions, and Depth Transformer infills the masked code stacks in parallel, by autoregressively predicting the $D$ codes at each position.

\paragraph{\emph{Bidirectional} Spatial Transformer} 
Given a masked sequence of code stacks $\bS_{\setminus \vecm}$, bidirectional spatial transformer first embeds $\bS_{\setminus \vecm}$ using the code embeddings of RQ-VAE as
\begin{equation}
    \vecu_n = \mathrm{PE}_N(n) + 
    \begin{cases}
        \sum_{d=1}^D \vece(\bS_{nd}) & \text{if $\vecm_n = 0$} \\
        \vece_{\masktok} & \text{if $\vecm_n = 1$}
    \end{cases},
\end{equation}
where $\mathrm{PE}_N(n)$ is an embedding for position $n$, and $\vece_{\masktok} \in \RR^{n_z}$ is an embedding for $\masktok$. 
Then, the bidirectional self-attention blocks, $f_{\theta}^{\mathrm{spatial}}$, extracts the context vector $\vech_n$ to predict $\bS_n$ as
\begin{equation}
    (\vech_1, \cdots, \vech_N) = 
    f_\theta^{\mathrm{spatial}}(\vecu_1, \cdots, \vecu_N).
\end{equation}

\paragraph{Depth Transformer} 
Depth transformer autoregressively predicts $\bS_n=(\bS_{n1}, \cdots, \bS_{nD})$ at a masked position. 
The input of depth transformer $(\vecv_{nd})_{d=1}^D$ is defined as 
\begin{equation}
    \vecv_{nd} = \mathrm{PE}_D(d) + 
    \begin{cases}
        \vech_n & \text{if $d=1$} \\
        \sum_{d'=1}^{d-1} \vece(\bS_{nd'}) & \text{if $d > 1$}
    \end{cases}
\end{equation}
where $\mathrm{PE}_D(d)$ is the positional embedding for depth $d$.
Then, depth transformer $f_\theta^{\mathrm{depth}}$, which consists of causal attention blocks, outputs the logit $\vecp_{nd}$ to predict $\bS_{nd}$ as
\begin{equation}
    \label{eq:logit_softmax}
    \vecp_{nd} = f_\theta^{\mathrm{depth}}(\vecv_{n1}, \cdots, \vecv_{nd})\quad\text{and}\quad
    p_\theta(\bS_{nd}=k | \bS_{n,<d}, \bS_{\setminus \vecm}) = \mathrm{softmax}(\vecp_{nd})_k.
\end{equation}

We remark that the architecture of Contextual RQ-Transformer subsumes bidirectional transformers. Specifically, RQ-Transformer with $D=1$ is equivalent to a bidirectional transformer since the depth transformer becomes a multilayer perceptron with layer normalization~\cite{layernorm}.

\subsubsection{Training of Contextual RQ-Transformer} \label{sec:training_part}
For the training of Contextual RQ-Transformer, let us define a mask distribution $q(\vecm)$ with a mask scheduling function $\gamma$. Following previous approaches~\cite{MaskGIT,mask-predict,ShowMeWhat}, $\gamma$ is chosen to be decreasing and to satisfy $\gamma(0)=1$ and $\gamma(1)=0$. Then, $\vecm \sim q(\vecm)$ is specified as
\begin{equation} \label{eq:mask_fn}
    r \sim \mathrm{Unif}([0, 1))
    \quad \text{and} \quad
    \vecm \sim \mathrm{Unif}(\{\vecm : |\vecm| = \lceil \gamma(r) \cdot N\rceil\}),
\end{equation}
where $|\vecm| = \sum_{n \in [N]} \vecm_n$ is the count of masked positions.
Finally, the training objective of \ARmodel is to minimize the negative log-likelihood of masked code stacks: 
\begin{equation}
\label{eq:nll_loss}
    \cL = 
    \mathbb{E}_{\vecm \sim q(\vecm)}
    \left[
    \mathbb{E}_{\bS}
    \left[
        \sum_{n: \vecm_n=1} \sum_{d=1}^D -\log p_{\theta}(\bS_{nd} | \bS_{n,<d}, \bS_{\setminus \vecm})
    \right]
    \right].
\end{equation}


\subsection{Draft-and-Revise: Two-Phase Decoding with Global Contexts of Generated Imaegs} 
\label{sec:draft_and_revise}

We propose a decoding algorithm, \emph{Draft-and-Revise}, which uses Contextual RQ-Transformer to effectively generate high-quality images with diverse visual contents. 
We introduce the details of Draft-and-Revise decoding and then explain how the two-phase decoding can effectively control the quality-diversity trade-off of generated images.

\begin{algorithm}[t]
\caption{\textsc{UPDATE} of $\bS$}
\label{alg:UPDATE}
\begin{algorithmic}[1]
\Require A sequence of code stacks $\bS$, a partition $\bPi = (\vecm^1, \cdots, \vecm^T)$, a model $\theta$
\For{$t=1, \cdots, T$}
    \State Sample $\bS_n \sim p_\theta (\bS_n | \bS_{\setminus \vecm^t})$ \; $\forall n: \vecm^t_n=1$
    \Comment{update the codes at masked positions}
\EndFor
\State \Return $\bS$
\end{algorithmic}
\end{algorithm}

\begin{algorithm}[t]
\caption{Draft-and-Revise decoding}
\label{alg:draft_then_revise}
\begin{algorithmic}[1]
\Require Partition sampling distributions $p_{\text{draft}}$ and $p_{\text{rev}}$, the number of revision iterations $M$
\item[] {\color{gray} \texttt{/* draft phase */}}
\State $\bS^{\text{empty}} \gets (\masktok, \cdots, \masktok)^N$
\Comment{initialize empty code map}
\State Sample $\bPi \sim p(\bPi;T_\text{draft})$
\State  
$\bS^{\text{draft}} \gets \textsc{UPDATE}(\bS^{\text{empty}}, \bPi; \theta)$
\Comment{generate a draft code map}
\item[] {\color{gray} \texttt{/* revision phase */}}
\State $\bS^{0} \gets \bS^{\text{draft}}$
\For{$m=1, \cdots, M$}
    \State Sample $\bPi \sim p(\bPi;T_\text{revise})$
    \State $\bS^m \gets \textsc{UPDATE}(\bS^{m-1}, \bPi; \theta)$
    \Comment{iteratively revise the code map}
\EndFor
\State \Return $\bS^M$
\end{algorithmic}
\end{algorithm}

We define a partition $\bPi = (\vecm^1, \cdots, \vecm^T)$ as a collection of pairwise disjoint $T$ mask vectors to cover all spatial positions, where $\sum_{t=1}^T \vecm^t_n = 1$ for all $n \in [N]$.
A partition $\bPi$ is sampled from the distribution $p(\bPi; T)$, which is the uniform distribution over all balanced partitions with size $T$:
\begin{equation}
    p(\bPi;T) = 
    \mathrm{Unif}\left(\{
        \bPi=(\vecm^1, \cdots, \vecm^T) : 
        |\vecm^t| = {\textstyle \frac{N}{T}} \ \,\forall t \in [T]
    \}\right).
\end{equation}

We first define a procedure $\textsc{UPDATE}(\bS, \bPi)$ to update $\bS$ as described in Algorithm~\ref{alg:UPDATE}, which updates $\bS_n$ with $\vecm_n^t=1$ for $t \in [T]$.
Then, \emph{Draft-and-Revise} decoding in Algorithm~\ref{alg:draft_then_revise} generates a draft from the empty sequence of code stacks and improves the quality of the draft.

\paragraph{Draft phase} 
In the draft phase, our model gradually infills the empty sequence of code stacks to generate a draft image, considering the global contexts of infilled code stacks.
Let $\bS^{\text{empty}}$ be an empty sequence of code stacks with $\bS^{\text{empty}}_{n} = \masktok^D$ for all $n$. Given a partition size $T_{\text{draft}}$, our model generates a draft image as 
\begin{equation}
    \bS^{\textrm{draft}} = \textsc{UPDATE}(\bS^{\textrm{empty}}, \bPi; \theta)
    \quad\text{where}\quad
    \bPi \sim p(\bPi; T_\text{draft}).
\end{equation}

\paragraph{Revise phase} 
The generated draft $\bS^{\text{draft}}$ is repeatedly revised to improve the visual quality of the image, while preserving the overall structure of the draft.
Given a partition size $T_{\text{revise}}$ and the number of updates $M$, the draft $\bS^0 = \bS^\text{draft}$ is repeatedly updated $M$ times as 
\begin{equation}
    \bS^{m} = \textsc{UPDATE}(\bS^{m-1}, \bPi; \theta)
    \quad\text{where}\quad
    \bPi \sim p(\bPi; T_\text{revise})
    \quad\text{for $m=1,\cdots,M$.}
\end{equation}

Note that Draft-and-Revise is not a tailored method, since we can adopt any mask-infilling-based generation method~\cite{UnleashingTransformer,MaskGIT} for \textsc{UPDATE} in Algorithm~\ref{alg:UPDATE}.
For example, confidence-based decoding~\cite{MaskGIT,ShowMeWhat}, which iteratively updates $\bS$ from high-confidence to low-confidence predictions, can be used for \textsc{UPDATE}.
However, we find that confidence-based decoding generates low-diversity images with oversimplified contents, since a model tends to predict simple visual patterns with high confidence.
In addition, confidence-based decoding often leads to biased unmasking patterns, which are not used in training, as shown in Appendix.
Thus, we use a uniformly random partition $\bPi$ in \textsc{UPDATE} as the most simplified rule, leaving investigations on sophisticated update methods as future work.


We postulate that our Draft-and-Revise can generate high-quality images with diverse contents by explicitly dividing two phases.
Specifically, a model first generates draft images with diverse visual contents despite the rather low quality of drafts.
After semantically diverse images are generated as drafts, we use sampling strategies such as temperature scaling~\cite{hinton2015distilling} and classifier-free guidance~\cite{ho2021classifierfree} in the revise phase to improve the visual quality of the drafts, while preserving the major semantic contents in drafts.
Thus, our method can improve the performance of image generation by effectively controlling the quality-diversity trade-off.
In addition, we emphasize that the two-phased decoding is intuitive and resembles the image generation process of human experts, who repeatedly refine their works to improve the quality after determining the overall contents first.

\begin{table} 
\centering
\small
\caption{FIDs, ISs, Precisions, and Recalls for class-conditional generation on ImageNet~\cite{deng2009imagenet}. $\dagger$ denotes the use of pretrained classifier for rejection sampling, gradient guidance, or training.}
\label{tab:result_cIN}
\begin{tabular}{l|c|c|cccc}
\toprule 
  & Params & $H\times W \times D$ & FID$\downarrow$ & IS$\uparrow$ & Precision$\uparrow$ & Recall$\uparrow$ \\ \hline
BigGAN-deep~\cite{BigGAN} & 112M & - & 6.95 & 202.6 & 0.87 & 0.23  \\ 
StyleGAN-XL$^\dagger$~\cite{styleganXL} & 166M & - & 2.3 & 262.1 & 0.78 & 0.53 \\ \hline
ADM~\cite{ADM} & 554M & - & 10.94 & 101.0 & 0.69 & 0.63\\ 
ADM-G$^\dagger$~\cite{ADM} & 608M & - & 4.59 & 186.7 & 0.82 & 0.52 \\ 
ImageBART~\cite{ImageBART} & 3.5B & 16$\times$16$\times$1 & 21.19 & 61.6 & - & -\\ 
VQ-Diffusion~\cite{VQDiff} & 518M & 16$\times$16$\times$1 & 11.89 & - & - & -  \\ 
LDM-8~\cite{LDM} & 395M & 32$\times$32 & 15.51 & 79.03 & 0.65 & 0.63 \\
LDM-8-G$^\dagger$~\cite{LDM} & 506M & 32$\times$32 & 7.76 & 209.52 & 0.84 & 0.35 \\
MaskGIT~\cite{MaskGIT} & 227M & 16$\times$16$\times$1 & 6.18 & 182.1 & 0.80 & 0.51 \\ \hline
VQ-GAN~\cite{VQGAN} & 1.4B & 16$\times$16$\times$1 & 15.78 & 74.3 & - & - \\ 
RQ-Transformer~\cite{RQVAE} & 1.4B & 8$\times$8$\times$4 & 8.71 & 119.0 & 0.71 & 0.58 \\ 
RQ-Transformer~\cite{RQVAE} & 3.8B & 8$\times$8$\times$4 & 7.55 & 134.0 & 0.73 & 0.58 \\
RQ-Transformer$^\dagger$~\cite{RQVAE} & 3.8B & 8$\times$8$\times$4 & 3.80 & 323.7 & 0.82 & 0.50 \\ \hline
\textbf{\ARmodel} & 371M & 8$\times$8$\times$4 & 5.45 & 172.6 & 0.81 & 0.49\\
\textbf{\ARmodel} & 821M & 8$\times$8$\times$4 & 3.45 & 221.9 & 0.82 & 0.52\\
\textbf{\ARmodel} & 1.4B & 8$\times$8$\times$4 & 3.41 & 224.6 & 0.79 & 0.54\\ \hline
Validation Data & - & - & 1.62 & 234.0 & 0.75 & 0.67 \\ 
\bottomrule
\end{tabular}
\end{table}

\section{Experiments} \label{sec:experiments}
In this section, we show that our Draft-and-Revise with \ARmodel can outperform previous approaches for class- and text-conditional image generation.
In addition, we conduct an extensive ablation study to understand the effects of Draft-and-Revise decoding on the quality and diversity of generated images, and the sampling speed.
We use the publicly released RQ-VAE~\cite{RQVAE} to represent a 256$\times$256 resolution of images as 8$\times$8$\times$4 codes.
For a fair comparison, we make \ARmodel have the same model size as the previous RQ-Transformer~\cite{RQVAE}.
For training, the quarter-period of cosine is used as the mask scheduling function $\gamma$ in Eq.~\ref{eq:mask_fn} following the previous studies~\cite{MaskGIT,improvedDDPM}.
We include the implementation details in Appendix.

\begin{figure}
\centering
\includegraphics[width=\textwidth]{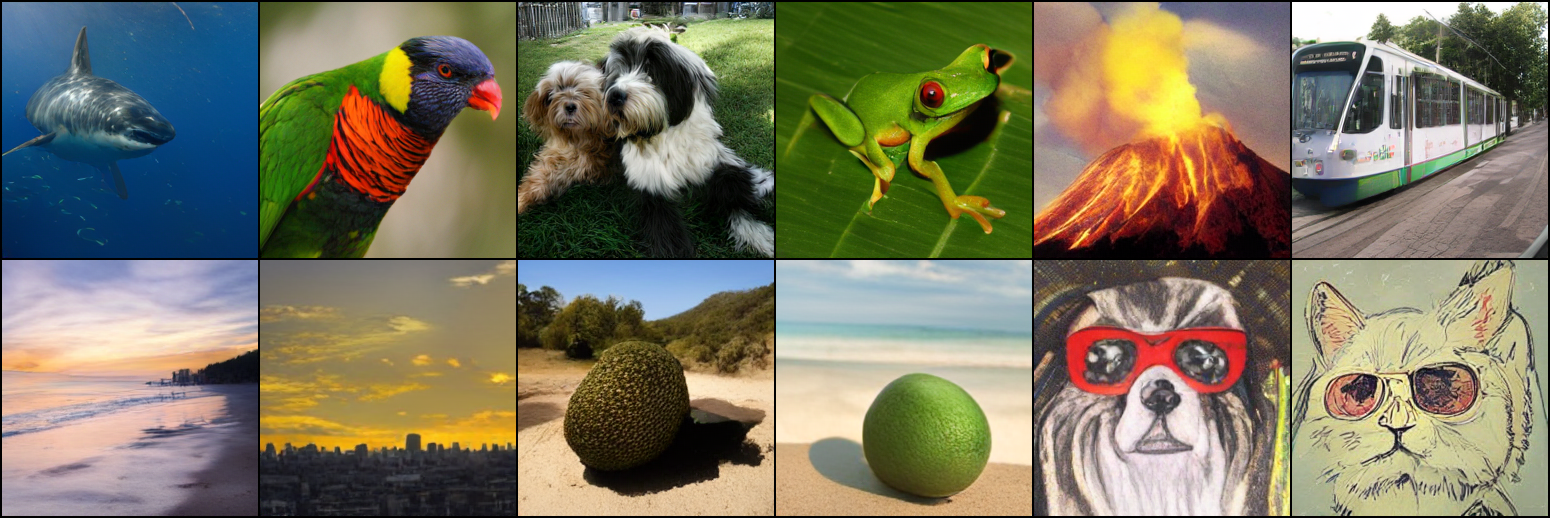}
\caption{The examples of generated 256$\times$256 images of our model trained on (Top) ImageNet and (Bottom) CC-3M. The used text conditions are "Sunset over the skyline of a \{beach, city\}.", "an avocado \{in the desert, on the seashore\}.", and "a painting of a \{dog, cat\} with sunglasses.".}
\label{fig:generated_images}
\end{figure}

\subsection{Class-conditional Image Generation} \label{sec:exp_class}

We train \ARmodel with 371M, 821M, and 1.4B parameters on ImageNet~\cite{deng2009imagenet} for class-conditional image generation.
For Draft-and-Revise decoding, we use $T_\text{draft}=64$, $T_\text{revise}=2$, and $M=2$.
We use temperature scaling~\cite{hinton2015distilling} and classifier-free guidance~\cite{ho2021classifierfree} only in the revise phase, while none of the strategies are applied in the draft phase.
Fréchet Inception Distance (FID)~\cite{heusel2017gans}, Inception Score (IS)~\cite{inception-score}, and Precision and Recall~\cite{precision-recall} are used for evaluation measures.

Table~\ref{tab:result_cIN} shows that \ARmodel significantly outperforms the previous approaches.
Notably, \ARmodel with 371M parameters outperforms RQ-Transformers with 1.4B and 3.8B parameters on all evaluation measures, despite having only about 4.2$\times$ and 11.4$\times$ fewer parameters.
In addition, the performance is improved as the number of parameters increases to 821M and 1.4B.
\ARmodel can achieve the lower FID score without a pretrained classifier than ADM-G and 3.8B parameters of RQ-Transformer with the use of pretrained classifier. StyleGAN-XL also uses a pretrained classifier during both training and image generation and achieves the lowest FID in Table~\ref{tab:result_cIN}.
However, our model with 1.4B parameters has higher precision and recall than StyleGAN-XL, implying that our model generates images of better fidelity and diversity without a pretrained classifier.
Our high performance without a classifier is remarkable, since the gradient guidance and rejection sampling are the tailored techniques to the model-based evaluation metrics in Table~\ref{tab:result_cIN}.
Considering that the performance is marginally improved as the number of parameters increases from 821M to 1.4B, an improved RQ-VAE can boost the performance of Contextual RQ-Transformer, since the reconstruction quality determines the best results of generated images.

\begin{wraptable}[11]{r}{0.51\textwidth}
\centering
\small
\caption{FIDs and CLIP scores~\cite{CLIP} on the validation dataset of CC-3M~\cite{sharma-etal-2018-conceptual} for T2I generation.}
\label{tab:exp_cc}
\begin{tabular}{l|c|ccc}
\toprule 
     & Params & FID$\downarrow$ & CLIP-s$\uparrow$ \\ \hline
VQ-GAN~\cite{VQGAN} & 600M & 28.86 &  0.20 \\ 
ImageBART~\cite{ImageBART} & 2.8B & 22.61 & 0.23 \\ 
LDM-4~\cite{LDM} & 645M & 17.01 & 0.24 \\
RQ-Transformer~\cite{RQVAE} & 654M & 12.33 & 0.26  \\ \hline
\textbf{Ours} & 366M & 10.44 & 0.26  \\ 
\textbf{Ours} & 654M & 9.65 & 0.26 \\ 
\bottomrule
\end{tabular}
\end{wraptable}

\subsection{Text-conditional Image Generation} \label{sec:exp_text}
We train \ARmodel with 366M and 654M parameters on CC-3M~\cite{sharma-etal-2018-conceptual} for text-to-image (T2I) generation.
We use Byte Pair Encoding~\cite{sennrich2016neural,huggingface} to encode a text condition into 32 tokens.
We also report CLIP-score~\cite{CLIP} with ViT-B/32~\cite{ViT} to measure the correspondence between texts and images.

\ARmodel in Table~\ref{tab:exp_cc} outperforms the previous T2I generation models.
\ARmodel with 366M parameters achieves better FID than RQ-Transformer with 654M parameters, and outperforms ImageBART and LDM-4, although our model has 12$\times$ fewer parameters than ImageBART.
When we increase the number of parameters to 654M, our model achieves state-of-the-art FID on CC-3M.
Meanwhile, our model does not improve the CLIP score of RQ-Transformer, but achieves competitive results with fewer parameters.
In Figure~\ref{fig:generated_images}, our model generates images with unseen texts in CC-3M.


\subsection{Conditional Image Inpainting} \label{sec:exp_manipulation}
We conduct conditional image inpainting where a model infills a masked area according to the given condition and contexts.
Figure~\ref{fig:inpainting} shows the example of image inpainting by RQ-Transformer (middle) and \ARmodel (right), when the class-condition is \emph{school bus}.
RQ-Transformer cannot attend to the right and bottom sides of the masked area and fails to generate a coherent image with given contexts. 
However, our model can complete the image to be coherent with given contexts by exploiting global contexts.
We attach more examples of image inpainting in Appendix.

\begin{figure}
\centering
\subfloat[]{\includegraphics[height=0.14\textheight]{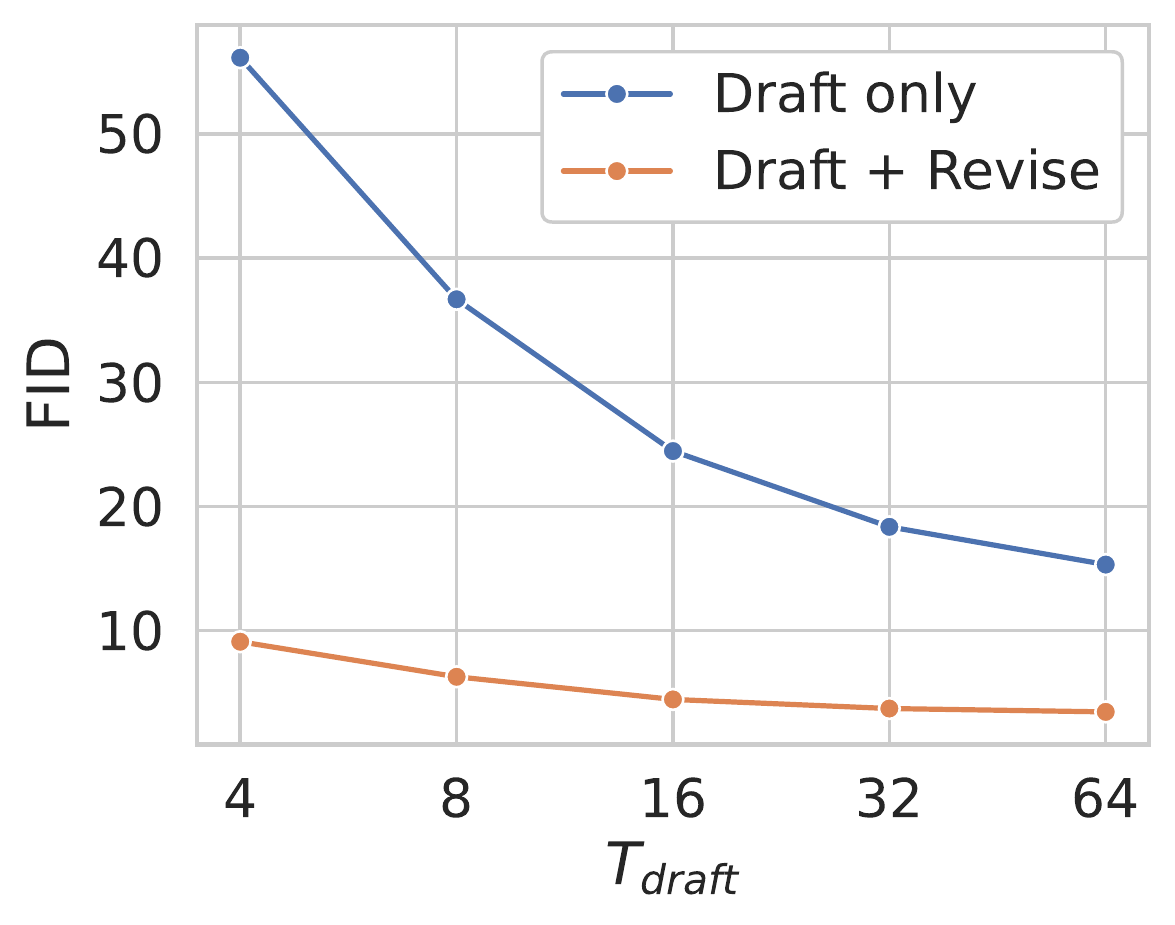}}
\hspace{4pt}
\subfloat[]{\includegraphics[height=0.14\textheight]{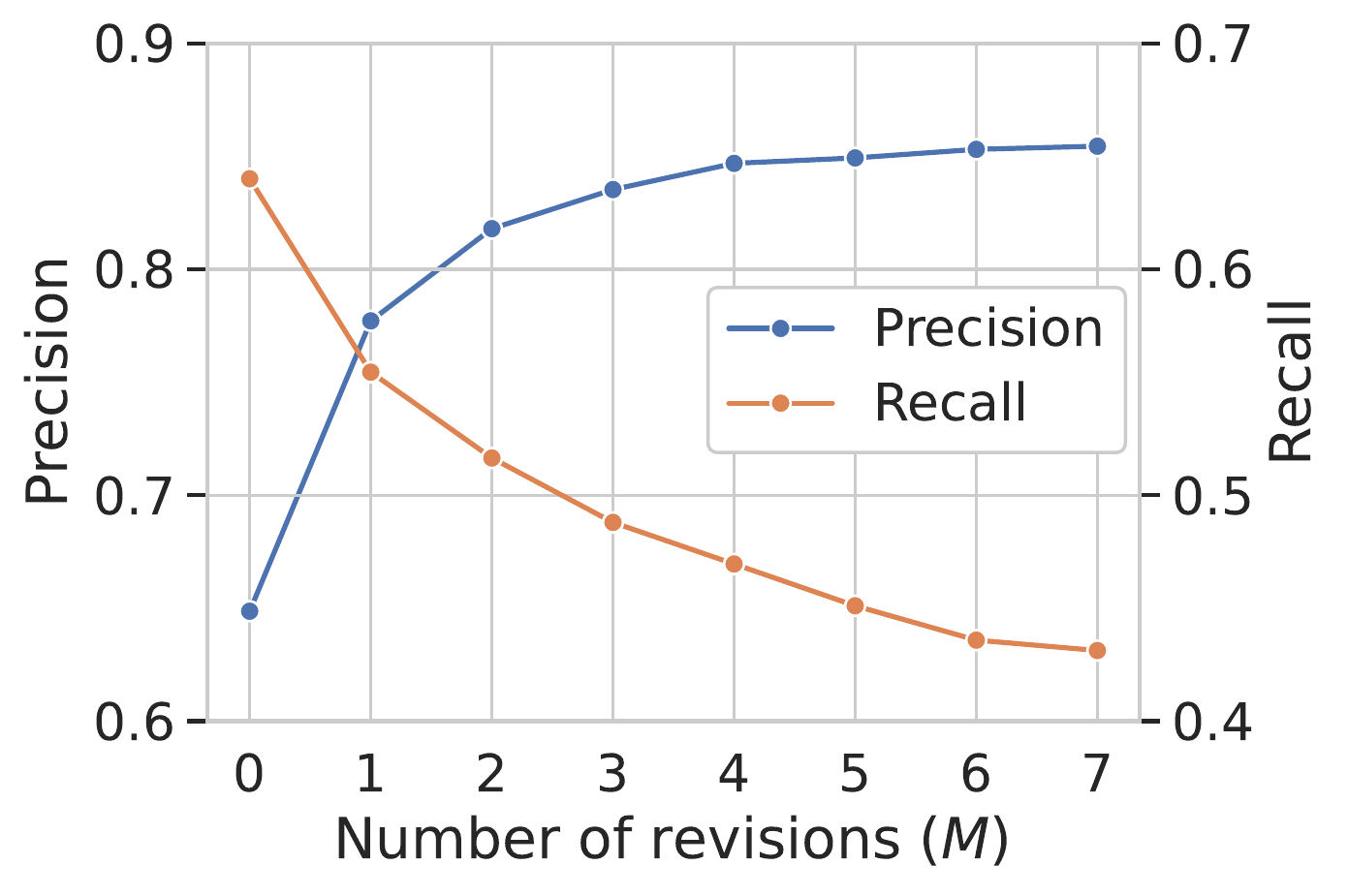}}
\hspace{3.5pt}
\subfloat[]{\includegraphics[height=0.14\textheight]{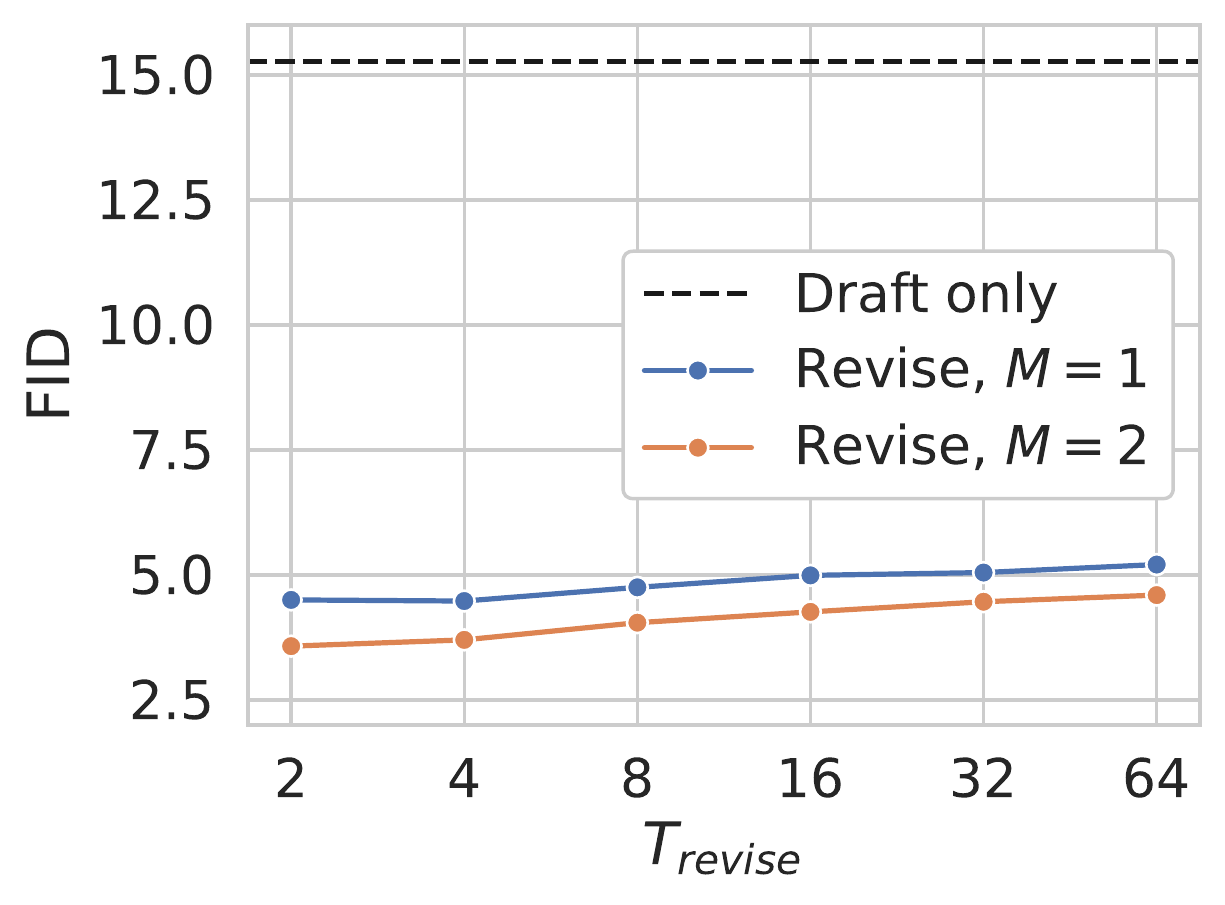}}
\caption{Ablation study on Draft-and-Revise decoding in Section~\ref{sec:exp_abl}. (a) FID subject to $T_{\text{draft}}$. (b) Precision and recall subject to $M$. (c) FID subject to $T_{\text{revise}}$. }
\label{fig:exp_abl}
\end{figure}

\begin{figure}
\centering
\includegraphics[width=\textwidth]{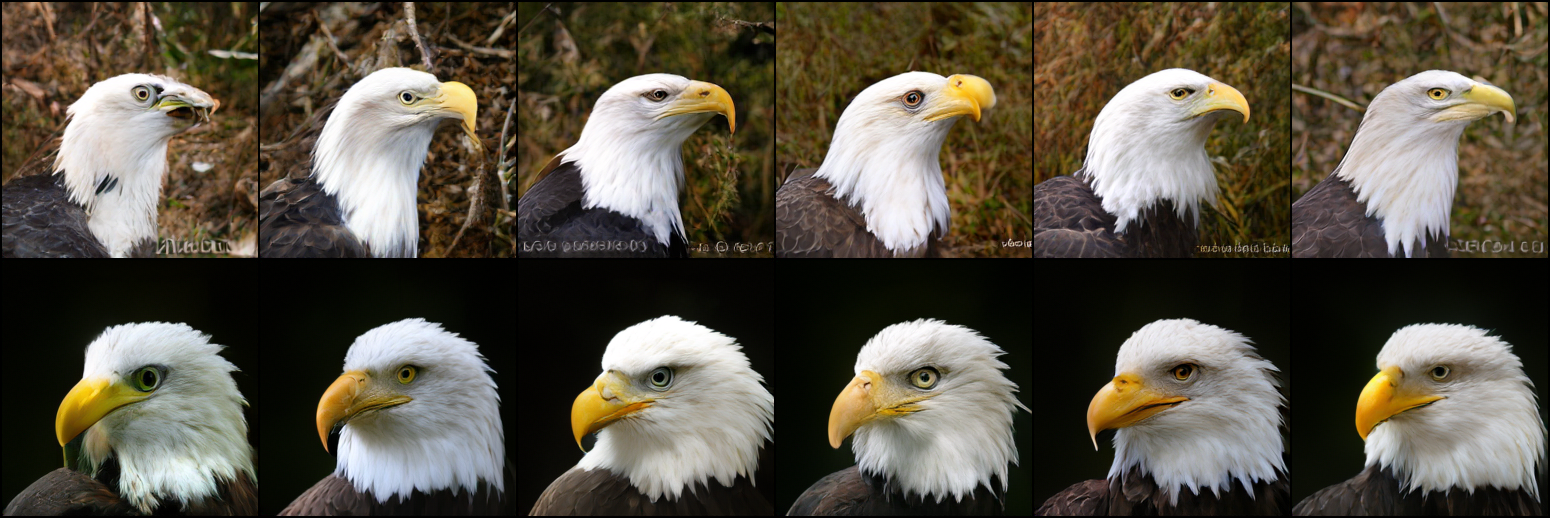}
\caption{Examples of generated images in the draft phase (left) and revise phases at $M=1,2,3,4,5$. The draft images are generated with $T_{\text{draft}}=8$ (top) and $T_{\text{draft}}=64$ (bottom), respectively.}
\label{fig:exp_revise_example}
\end{figure}

\subsection{Ablation Study on Draft-and-Revise} \label{sec:exp_abl}
We conduct an extensive ablation study to demonstrate the effectiveness of Draft-and-Revise decoding of our framework. We use \ARmodel with 821M parameters trained on ImageNet.

\paragraph{Quality improvement of draft images in the revise phase}
Figure~\ref{fig:exp_abl}(a) shows the effects of $T_\text{draft}$ on draft images and their quality improvement in the revised phase with  $T_\text{revise}=2$ and $M=2$.
In the draft phase, FID is improved as $T_\text{draft}$ increases from 4 to 64.
At each inference, \ARmodel generates $N/T_\text{draft}$ code stacks in parallel, starting with the empty sequence.
Thus, the model with a large $T_\text{draft}$ generates a small number of code stacks at each inference and can avoid generating incoherent code stacks in the early stage of the draft phase.
Although FIDs in the draft phase are worse, they are significantly improved in the revise phase as shown in Figure~\ref{fig:exp_revise_example}.


\paragraph{Effect of $M$ and $T_\text{revise}$ in the revise phase} 
Figure~\ref{fig:exp_abl}(b) shows the effects of the number of updates $M$ in the revise phase on the quality and diversity of generated images.
Since the quality-diversity trade-off exists as the updates are repeated, we select $M=2$ as the default hyperparameter to balance the precision and recall, considering that the increase of precision starts to slow down.
Interestingly, Figure~\ref{fig:exp_revise_example} shows that the overall contents remain unchanged even after $M>2$.
Thus, we claim that Draft-and-Revise decoding does not harm the perceptual diversity of generated images throughout the revise phase despite the consistent deterioration of recall.

Figure~\ref{fig:exp_abl}(c) shows the effects of $T_\text{revise}$ on the quality of generated images.
The FIDs are significantly improved in the revise phase regardless of the choice of $T_\text{revise}$, but increasing $T_\text{revise}$ slightly deteriorates FIDs.
We remark that some code stacks of a draft can be erroneous due to its low quality, and a model with large $T_\text{revise}$ slowly updates a small number of code stacks at once in the revise phase.  
Therefore, the updates with large $T_\text{revise}$ can be more influenced by the erroneous code stacks.
Although $T_\text{revise}=2$ updates half of an image at once, our draft-and-revise decoding successfully improves the quality of generated images, while preserving the global contexts of drafts, as shown in Figure~\ref{fig:exp_revise_example}.
The study on self-supervised learning~\cite{MAE} also reports similar results, where a masked auto-encoder reconstructs the global contexts of an image after masking half of the image.

\begin{wraptable}{r}{0.4\textwidth}
\small
\centering
\caption{The effects of classifier-free guidance on the image generation.}
\label{tab:exp_gg}
\begin{tabular}{cc|ccc}
\toprule
Draft & Revise & FID & P & R \\ \hline
 & & 5.78 & 0.72 & 0.58 \\
 & \checkmark & 3.45 & 0.82 & 0.52 \\
 \checkmark & \checkmark & 8.90 & 0.92 & 0.33 \\
\bottomrule
\end{tabular}
\end{wraptable}

\paragraph{Quality-diversity control of Draft-and-Revise} 
Our \emph{Draft-and-Revise} decoding can effectively control the quality-diversity trade-off in generated images.
Table~\ref{tab:exp_gg} shows FID, precision (P), and recall (R) according to the use of classifier-free guidance~\cite{ho2021classifierfree} with a scale of 1.8, while applying temperature scaling with 0.8 only to the revise phase.
\ARmodel without the guidance already outperforms RQ-Transformer with 3.8B parameters and demonstrates the effectiveness of our framework.
When the guidance is used for both draft and revise phases, the precision dramatically increases but the recall decreases to 0.33. 
Consequently, FID becomes worse due to the lack of diversity in generated images.
However, when the guidance is applied only to the revise phase, our model achieves the lowest FID, as the quality and diversity are well-balanced.
Thus, the explicitly separated two phases of Draft-and-Revise can effectively control the issue of quality-diversity trade-off by generating diverse drafts and then improving their quality.

\begin{wraptable}{r}{0.38\textwidth}
\small
\centering\caption{Comparison of FID and the sampling speed of image generation.}
\label{tab:speed}
\begin{tabular}{l|cc}
\toprule
  & FID & s/sample \\ \hline
 VQGAN & 15.78 & 0.16 \\
 RQ-Transformer & 8.71 & 0.04\\ \hline
 \rowcolor[gray]{0.95}\multicolumn{3}{c}{\textit{\textbf{Contextual RQ-Transformer}}} \\ \hline
 $T_{\text{draft}}=8$ & 5.41 & 0.03 \\
 $T_{\text{draft}}=32$ & 3.73 & 0.06 \\
 $T_{\text{draft}}=64$ & 3.45 & 0.10 \\
\bottomrule
\end{tabular}
\end{wraptable}

\paragraph{Trade-off between quality and sampling speed}
After we fix $T_\text{revise}=2$ and $M=2$, the trade-off between FID and the sampling speed is analyzed in Table~\ref{tab:speed} according to $T_\text{draft}$.
Following the previous study~\cite{RQVAE}, we generate 5,000 samples with batch size of 100.
\ARmodel with $T_\text{draft}=8$ outperforms VQGAN and RQ-Transformer with 1.4B parameters in terms of both FID and the sampling speed.
Although the sampling speed becomes slow with increased $T_\text{draft}$, the FID scores are consistently improved.
We remark that the sampling speed with $T_\text{draft}=64$ is 2.5$\times$ slower than RQ-Transformer, but our model outperforms 3.8B parameters of RQ-Transformer with rejection sampling in Table~\ref{tab:result_cIN}.
The results represent that our framework has inexpensive computational costs to generate high-quality images, since rejection sampling requires generating up to 20$\times$ more samples than ours.

\section{Conclusion} \label{sec:conclusion}
In this study, we have proposed \emph{Draft-and-Revise} for an effective image generation framework with Contextual RQ-Transformer.
After an image is represented as a sequence of code stacks, Contextual RQ-Transformer is trained to infill a randomly masked sequence.
Then, Draft-and-Revise decoding is used to generate high-quality images by first generating a draft image with diverse contents and then improving its visual quality based on the global contexts of the draft.
Consequently, we can achieve state-of-the-art results on ImageNet and CC-3M, demonstrating the effectiveness of our framework.

Our study has two main limitations to be further explored.
Firstly, Draft-and-Revise decoding always updates all code stacks in the revise phase, although some code stacks might not need an update.
In future work, a selective method can be developed to improve the efficiency of the revise phase by a sophisticated approach.
Secondly, our generative model is not validated on various downstream tasks.
Since masked token modeling is successful self-supervised learning for texts~\cite{BERT} and images~\cite{BeiT,MAE}, a unified model for both generative and discriminative tasks~\cite{GCRL} is worth exploration for future work.

\section{Acknowledgements}
This work was supported by Institute of Information \& communications Technology Planning \& Evaluation(IITP) grant funded by the Korea government(MSIT) (No.2018-0-01398: Development of a Conversational, Self-tuning DBMS; No.2021-0-00537: Visual Common Sense).

\bibliographystyle{plain}
\bibliography{neurips_2022}

\clearpage
\appendix


\section{Implementation Details}

\subsection{Details of RQ-VAE}

RQ-VAE~\cite{RQVAE} is a generalized version of VQ-VAE~\cite{VQVAE} and VQGAN~\cite{VQGAN}, since RQ-VAE with $D=1$ is equivalent to VQ-VAE or VQGAN.
When $D>1$, RQ-VAE recursively discretizes the feature map of an image for a precise approximation of the feature map using the codebook.
When the codebook size of RQ is $K$, RQ with depth $D$ is as capable as VQ with $K^D$ size of a codebook, since RQ can represent at most $K^D$ clusters in a vector space.
That is, if the codebook sizes are the same, RQ with $D>1$ can approximate a feature vector more accurately than VQ.
Thus, RQ-VAE can further reduce the spatial resolution of code map than VQ-VAE and VQ-GAN, and therefore outperforms previous autoregressive models with discrete representations.

Following the previous studies~\cite{VQGAN,RQVAE,VQVAE}, the training of RQ-VAE uses the reconstruction loss, the commitment loss, the adversarial training~\cite{isola2017image}, and the LPIPS perceptual loss~\cite{johnson2016perceptual}.
The codebook $\cC$ of RQ-VAE is updated using the exponential moving average during training~\cite{RQVAE,VQVAE}.

In experiments, we use the pretrained RQ-VAE, which is publicly available\footnote{\url{https://github.com/kakaobrain/rq-vae-transformer}}.
The RQ-VAE uses the codebook size of 16,384 to represent a 256$\times$256 resolution of an image as 8$\times$8$\times$4 shape of a code map.
The architecture of RQ-VAE is the same as VQGAN~\cite{VQGAN} except for adding residual blocks in the encoder and the decoder to reduce the spatial resolution of the code map more than VQGAN.

\subsection{Architecture of Contextual RQ-Transformer}

\begin{wraptable}{r}{0.5\textwidth}
\small
\centering\caption{Architecture details of Contextual RQ-Transformer for ImageNet and CC-3M.}
\label{tab:arch}
\begin{tabular}{ccccc}
\toprule
Dataset & \# params. & $N_\text{spatial}$ & $N_\text{depth}$ & $d_\text{model}$\\ \hline
\multirow{3}{*}{ImageNet} & 371M & 24 & 4 & 1024 \\
& 821M & 24 & 4 & 1536 \\
& 1.4B & 42 & 6 & 1536 \\ \hline
\multirow{2}{*}{CC-3M}& 366M & 21 & 4 & 1024 \\
& 654M & 26 & 4 & 1280 \\
\bottomrule
\end{tabular}
\end{wraptable}

Table \ref{tab:arch} summarizes the architecture details of Contextual RQ-Transformers to be trained on ImageNet and CC-3M. Contextual RQ-Transformer consists of two compartments: \emph{bidirectional spatial transformer} with $N_\text{spatial}$ self-attention blocks and \emph{depth transformer} with $N_\text{depth}$ causal self-attention blocks. 
The dimensionality of embeddings in multi-headed self-attentions is denoted $d_\text{model}$, while the dimensionality for each attention head is 64.

\subsection{Training details}

All Contextual RQ-Transformers are trained with AdamW optimizer with $\beta_1=0.9$, $\beta_2=0.95$, and weight decay 0.0001. 
Each model is trained for 300 epochs with the cosine learning rate schedule with the initial value of 0.0001 and the final value of 0, for both ImageNet and CC-3M. 
We use eight NVIDIA A100 GPUs to train the models with 821M and 1.4B parameters on ImageNet and the model with 650M parameters on CC-3M, while four GPUs are used for the models with 366M parameters.
For our model with 821M and 1.4B parameters on ImageNet, the training takes at most 10 days.

\subsection{Draft-and-Revise decoding details}
We use temperature scaling with the 0.8 scale in the revise phase and do not apply the temperature scaling in the draft phase.
The classifier-free guidance is also applied only to the sampling in the revise phase.
We use 1.4, 1.8, and 2.0 scales of guidance for 371M, 821M, and 1.4B parameters of Contextual RQ-Transformer on ImageNet, respectively.
In addition, $(M, T_\text{revise})$ is (3,2), (2,2), and (2,2), respectively.
In the revise phase for Contextual RQ-Transformer on CC-3M, we use the 1.1 scale of classifier-free guidance with $(M, T_\text{revise})=(2,4)$.

\section{The Compatibility of Draft-and-Revise with 16$\times$16 VQGAN}

\begin{table} 
\small
\centering\caption{Performance of Contextual RQ-Transformers on 16$\times$16$\times$1 RQ-VAE.}
\label{tab:rq_16x16}
\begin{tabular}{l|c|c|cccc}
\toprule
 & \# params & $H\times W\times D$ & FID & P & R & s/sample \\ \hline
VQ-GAN~\cite{VQGAN} & 1.4B & 16$\times$16$\times$1 & 15.78 & - & - & - \\ 
RQ-Transformer~\cite{RQVAE} & 1.4B & 8$\times$8$\times$4 & 8.71 & 0.71 & 0.58 & 0.04 \\ \hline

Contextual VQ-Transformer & 350M & 16$\times$16$\times$1 & 6.44 & 0.79 & 0.47 & 0.83 \\ \hline
Contextual RQ-Transformer & 371M & 8$\times$8$\times$4 & 5.45 & 0.81 & 0.49 & 0.08 \\
\bottomrule
\end{tabular}
\end{table} 

\begin{figure}
\centering
\subfloat[]{\includegraphics[height=0.2\textheight]{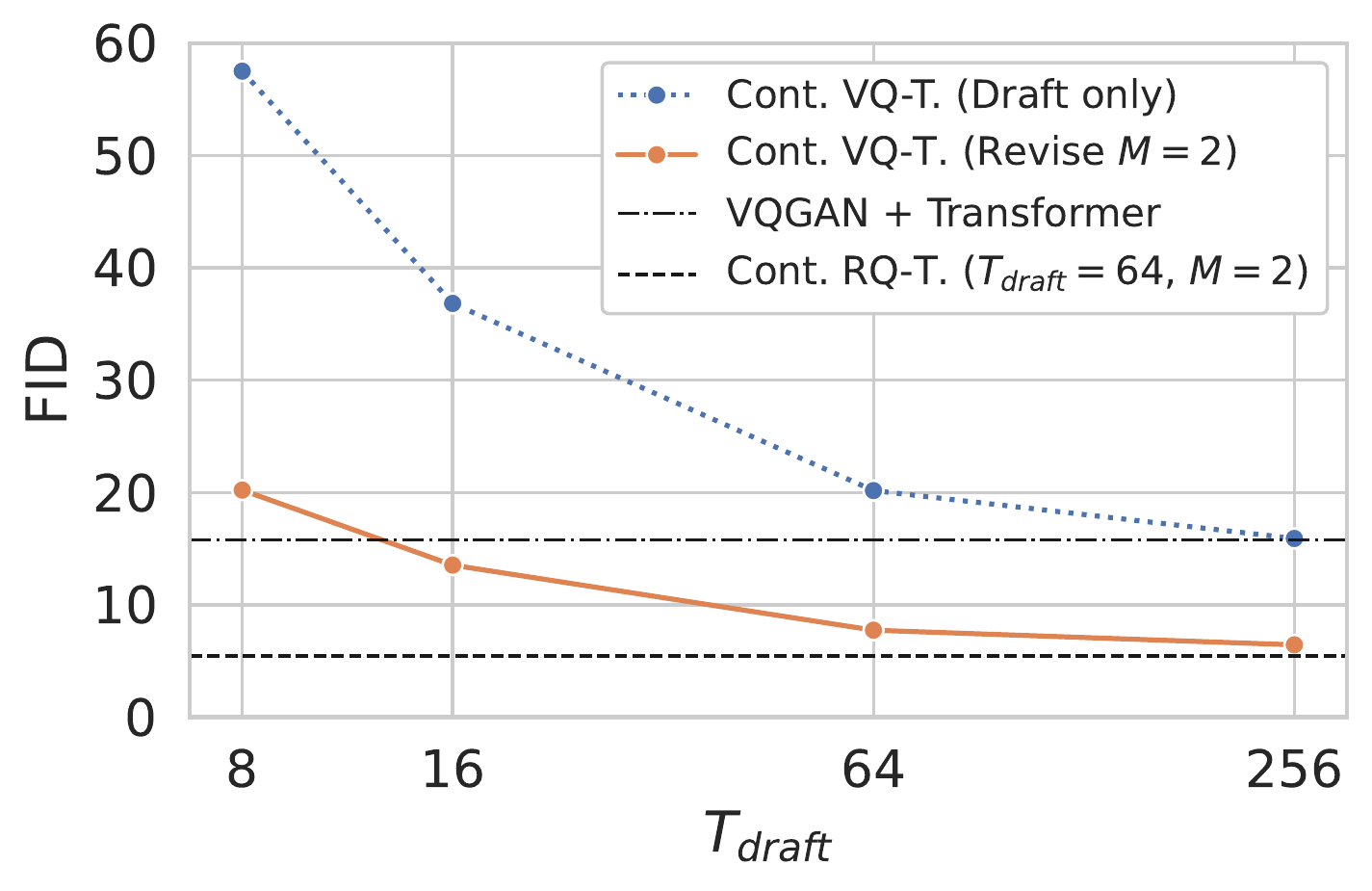}}
\subfloat[]{\includegraphics[height=0.2\textheight]{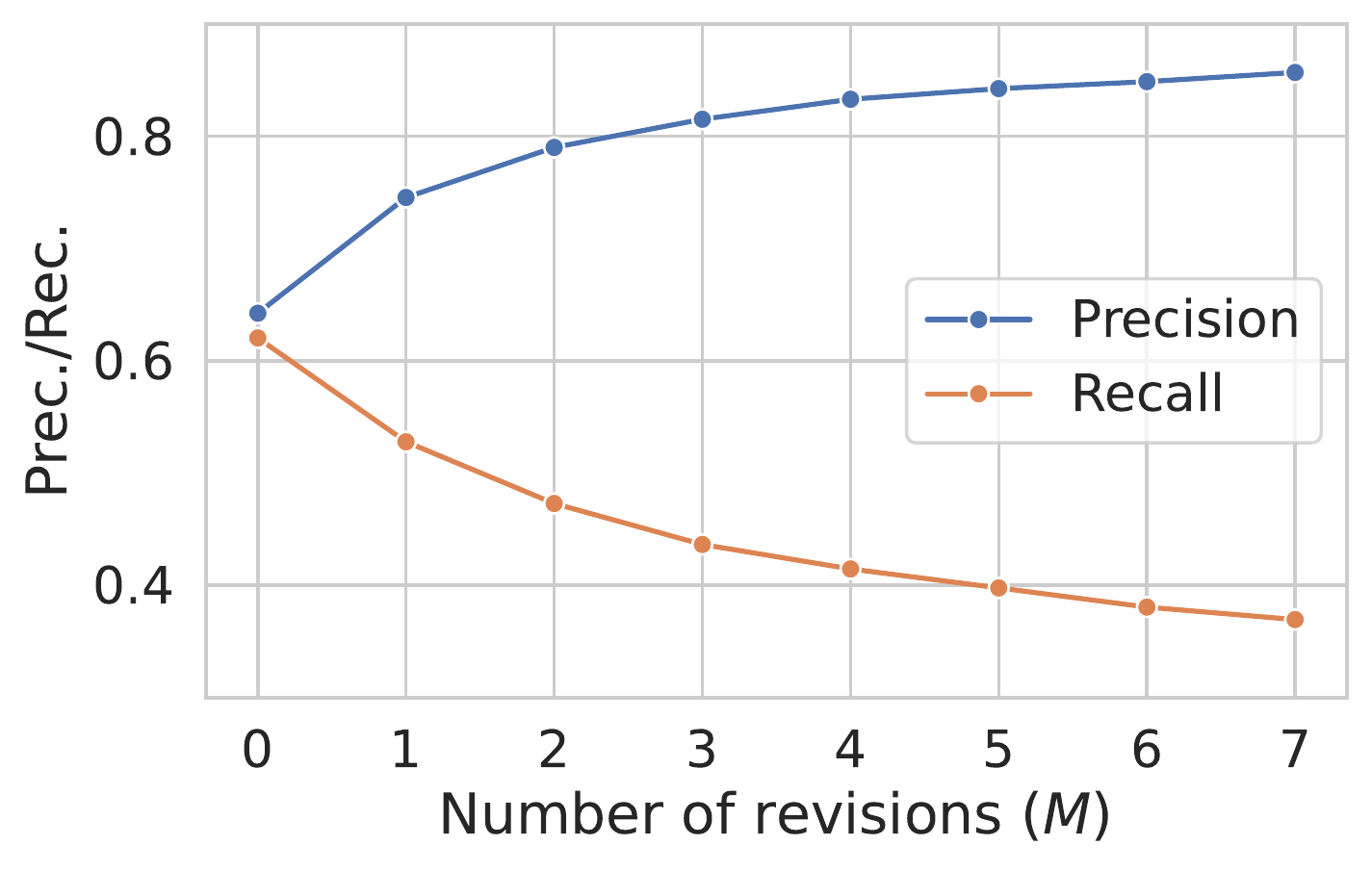}}
\caption{Effect of Draft-and-Revise decoding on Contextual VQ-Transformer. (a) FID subject to $T_{\text{draft}}$. (b) Precision and recall subject to $M$.}
\label{fig:rq_16x16}
\end{figure}

Since RQ is a generalized VQ, our framework of Draft-and-Revise with Contextual RQ-Transformer is also applicable to VQ-VAE~\cite{VQVAE} and VQGAN~\cite{VQGAN}.
Note that RQ-VAE with $D=1$ is equivalent to VQGAN, where their spatial resolutions of code maps are the same.
To validate that our framework is also effective on the 16$\times$16 shape of code map by VQGAN, we first train an RQ-VAE, which represents an image as 16$\times$16$\times$1 shape of code maps and has the identical architecture to VQGAN with 16$\times$16 shape of code map.
Then, we train a Contextual RQ-Transformer with 350M parameters on 16$\times$16$\times$1 codes. 
We notate the Contextual RQ-Transformer with $D=1$ as \emph{Contextual VQ-Transformer} throughout this section.

Figure \ref{fig:rq_16x16} shows that the Draft-and-Revise decoding also generalizes to Contextual VQ-Transformer and can control the quality-diversity trade-off in the same manner as with the Contextual RQ-Transformer. 
We fix $T_\text{revise}=2$ and $M=2$, and use temperature scaling of 0.8 and classifier-free guidance of 2.4 only in the revise phase. 
As shown in Figure \ref{fig:rq_16x16}(a), the best performance is achieved when $T_\text{draft}=256$ and the corresponding FID, precision (P), and recall (R) are 6.44, 0.79, and 0.47, respectively, as reported in Table \ref{tab:rq_16x16}.
Note that Contextual VQ-Transformer outperforms the AR model such as VQGAN, although our model has $4\times$ fewer parameters than the AR model.
In addition, our draft-and-revise decoding with Contextual VQ-Transformer also works as well as Contextual RQ-Transformer, showing that the iterative updates in the revise phase consistently increase precisions and decreases recalls in Figure~\ref{fig:rq_16x16}(b).
Consequently, the results validate that our framework is more effective for high-quality image generation than AR modeling.

Although our framework is compatible with VQGAN, we emphasize that Contextual RQ-Transformer is more effective than Contextual VQ-Transformer.
Contextual RQ-Transformer with a similar number of parameters outperforms the Contextual VQ-Transformer in terms of FID, precision, and recall in Table~\ref{tab:rq_16x16}.
In addition, Contextual RQ-Transformer has about 10$\times$ faster speed for image generation than Contextual VQ-Transformer, since the computational complexity of self-attention is mainly determined by the sequence length.
Although the comparison of FID, precision, and recall is not entirely fair due to the different performance between RQ-VAE and VQGAN, Contextual RQ-Transformer can achieve state-of-the-art performance with lower computational costs than Contextual VQ-Transformer.
Thus, masked modeling in RQ representations is more effective and efficient than in VQ representations, if equipped with our Contextual RQ-Transformer.


\section{Comparison with Confidence-based Mask-infilling Strategies}


\begin{table}[t] 
\small
\centering\caption{Comparison of confidence-based mask-infilling strategies and our random partitioning strategy in the draft phase.}
\label{tab:conf_vs_rand}
\begin{tabular}{l|ccc|ccc}
\toprule
& \multicolumn{3}{c|}{Draft Images} & \multicolumn{3}{c}{Revised Images} \\ \cline{2-7}
  & FID & P & R & FID & P & R \\ \hline
 \rowcolor[gray]{0.95}\multicolumn{7}{c}{$T_\text{draft}=8$} \\ \hline
 Top-C & 18.28 & 0.67 & 0.43 & 11.92 & 0.79 & 0.35 \\ 
 Top-C-50\% & 13.26 & 0.71 & 0.53 & 6.13 & 0.86 & 0.41 \\
 Random & 36.68 & 0.58 & 0.60 & 6.28 & 0.80 & 0.49 \\ \hline
 \rowcolor[gray]{0.95}\multicolumn{7}{c}{$T_\text{draft}=64$} \\ \hline
 Top-C & 34.07 & 0.54 & 0.42 & 27.88 & 0.62 & 0.37 \\
 Top-C-50\% & 7.22 & 0.75 & 0.53 & 6.84 & 0.84 & 0.42 \\
 Random & 15.32 & 0.65 & 0.64 & 3.45 & 0.82 & 0.52 \\ 
\bottomrule
\end{tabular}
\end{table} 

In this section, we examine how the selection of \textrm{UPDATE} in Algorithm 1 affects the performance of image generation.
First, instead of using our random updates of spatial positions, we consider a confidence-based mask-infilling strategy of MaskGIT~\cite{MaskGIT} and denote the sampling strategy of MaskGIT as \emph{Top-C}.
At each inference step, \emph{Top-C} considers the predicted confidences and determines the unmasked positions to have highly confident predictions.
Then, we also consider a mixed strategy, \emph{Top-C-50\%}, which first filters out the bottom 50\% confident positions, and then randomly selects the unmasked positions among the positions with top 50\% high confidence.
\emph{Top-C-50\%} is similar to the combination of random sampling after top-k or top-p~\cite{topKP} filtering.
We denote our mask-infilling strategy as \emph{Random}, which randomly determines the unmasked regions at each inference in the draft phase.
We use Contextual RQ-Transformer with 821M parameters and fix the parameters of Draft-and-Revise decoding to $T_\text{draft}=64$, $T_\text{revise}=2$, and $M=2$ and apply temperature scaling of 0.8 and classifier-free guidance of 1.8 in the revise phase. We report FID, precision (P), and recall (R) of the generated images in the draft and revise phases.


Table \ref{tab:conf_vs_rand} shows the effect of Top-C, Top-C-50\%, and Random in Draft-and-Revise decoding.
Regardless of the selection of mask-infilling strategies in the draft phase, our draft-and-revise decoding consistently improves the performance of image generation after the revise phase.
The results imply that our framework of Draft-and-Revise can be effectively generalized to various approaches of mask-infilling-based image generation.

When $T_\text{draft}=8$, the draft images of Top-C have better FID but worse recall than the draft images of Random. Nonetheless, the quality of the draft images is significantly improved with the revise phase and subsequently Random outperforms Top-C in all three metrics. Top-C-50\% achieves the best FID in both the draft and revised images, but the recall of Top-C-50\% is significantly lower than the recall of Random. 
When $T_\text{draft}=64$, the draft images of Top-C exhibit worse metrics than the draft images of Random. Although Top-C-50\% achieves lower FID than Top-C and Random in terms of draft images, Random outperforms Top-C and Top-C-50\% after the revise phase. This supports our claim that Draft-and-Revise decoding better controls quality-diversity trade-off when drafts are generated with maximal diversity.

\begin{figure}
\centering
\includegraphics[width=\textwidth]{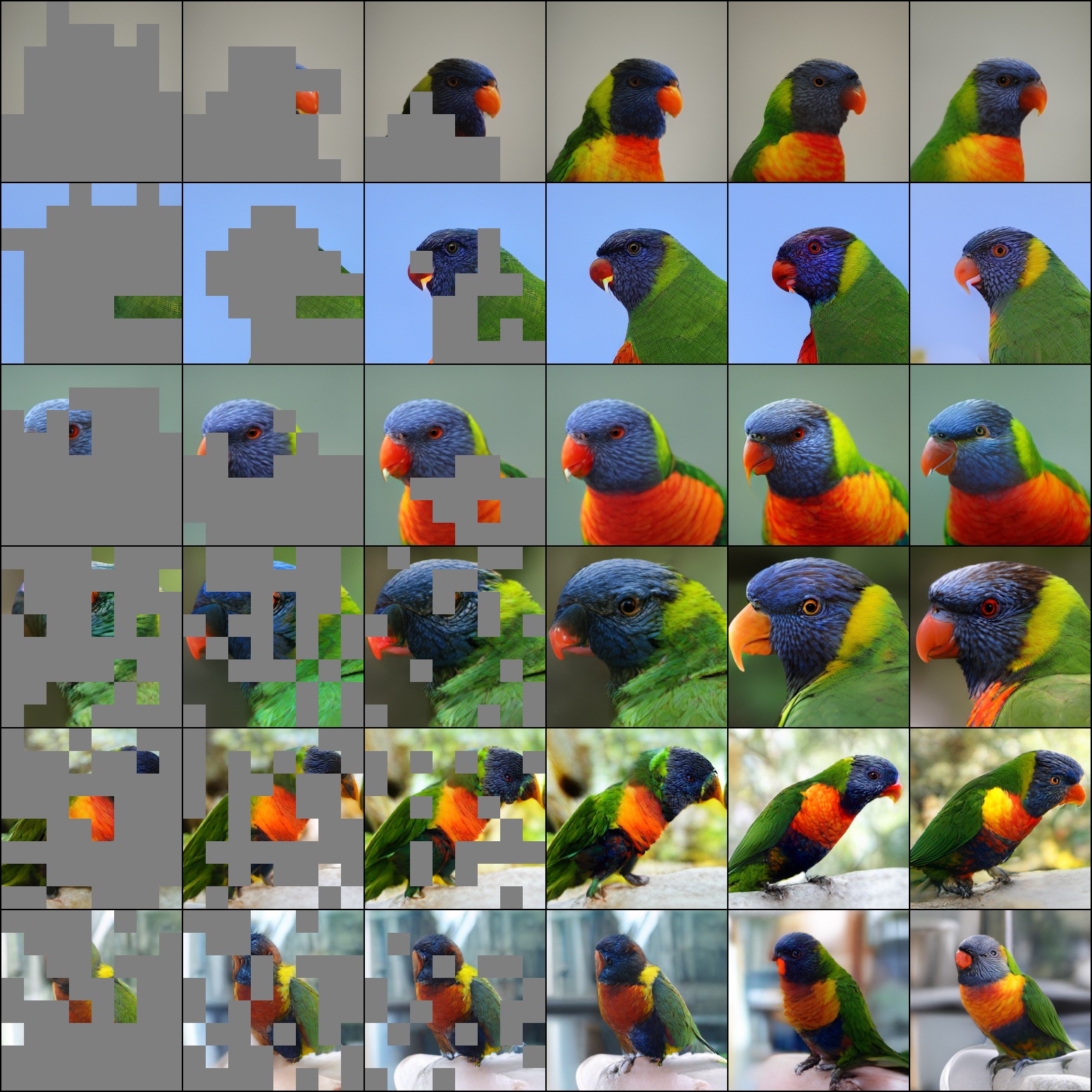}
\caption{Intermediary samples of mask-infilling strategies with $T_\text{draft}=64$ and subsequently revised samples. From left to right, the first four images are the samples with the corresponding masking pattern of every 16th mask-infilling step, and the last two images are the revised samples for $M=1,2$. (Top 3 rows) Mask-infilled with a confidence-based strategy, \emph{Top-C}. (Bottom 3 rows) Mask-infilled with our strategy, \emph{Random}.}
\label{fig:conf_vs_rand}
\end{figure}

By the visual analysis in Figure~\ref{fig:conf_vs_rand}, we find that the limited performance of confidence-based mask-infilling strategies results from the bias of a model on high-confident predictions.
That is, Top-C tends to predict only simple patterns in the early phase of mask-infilling, since the simple visual patterns are prone to have high-confident predictions. 
This effect becomes more apparent when $T_\text{draft}=64$ as only the code stack with the highest confidence is included in the sample at each inference, thereby the diversity of the generated images becomes severely limited. 
Indeed, it is shown in Figure \ref{fig:conf_vs_rand} that the confidence-based methods first infill the backgrounds with simple visual patterns, and the resulting samples exhibit low diversity of visual contents.
Due to the limited diversity, the confidence-based mask-infilling strategies limit the performance of FID, although our draft-and-revise decoding further improves the visual quality of generated images after the revise phase.
Thus, we conclude that our sampling strategy, which first generates diverse visual contents and then improves their quality, is effective for draft-and-revise decoding to achieve high performance of image generation.



\begin{figure}
    \centering
    \includegraphics[width=\textwidth]{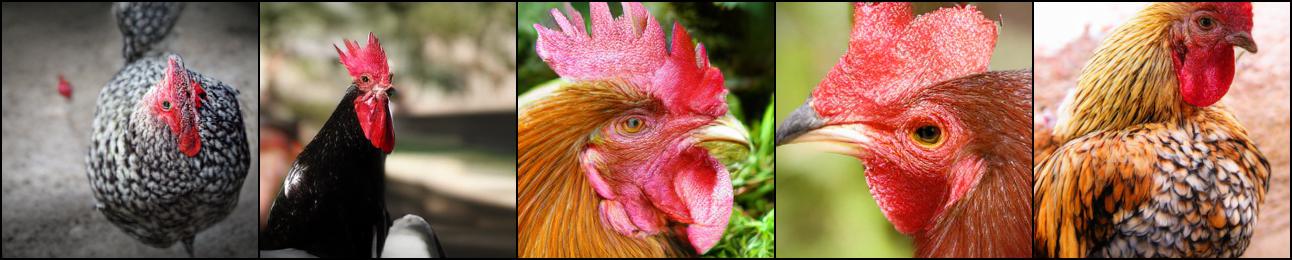}
    \includegraphics[width=\textwidth]{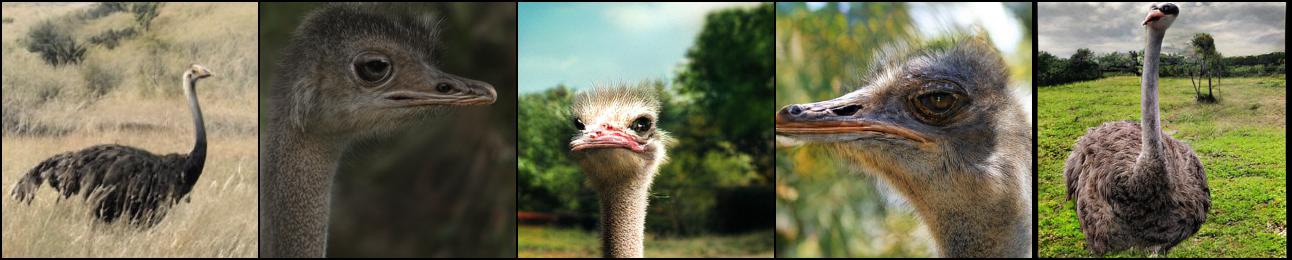}
    \includegraphics[width=\textwidth]{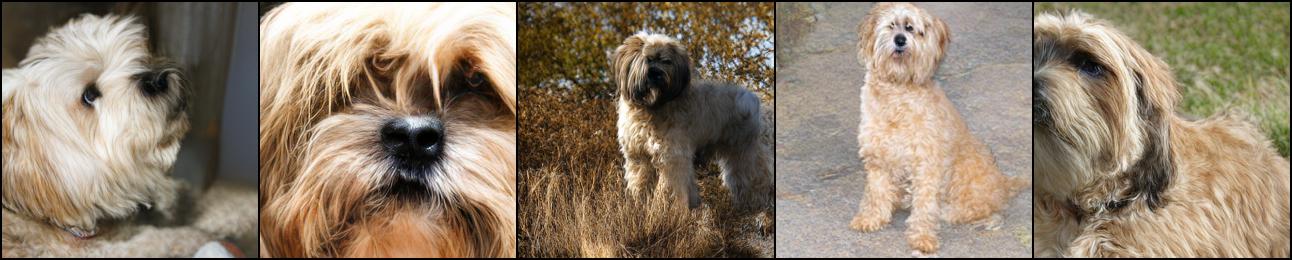}
    \includegraphics[width=\textwidth]{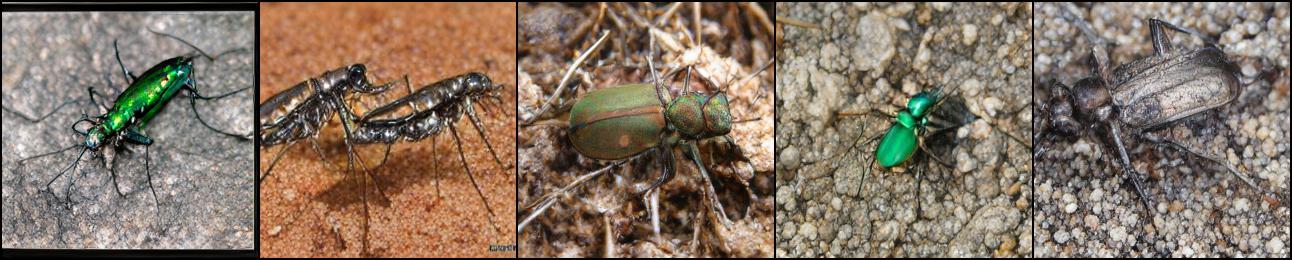}
    \includegraphics[width=\textwidth]{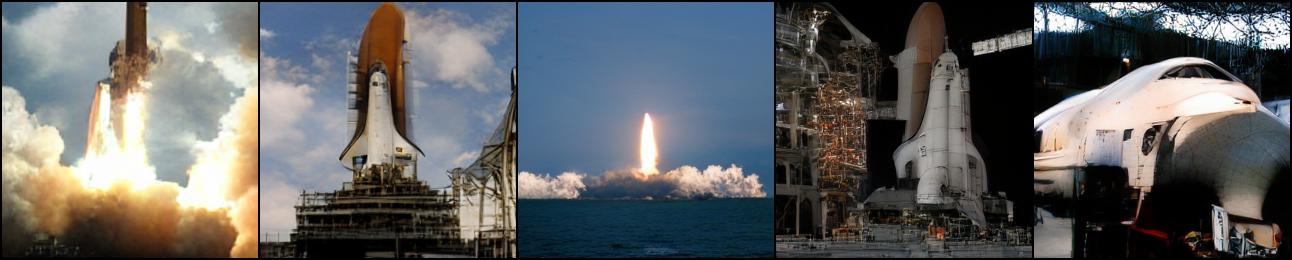}
    \includegraphics[width=\textwidth]{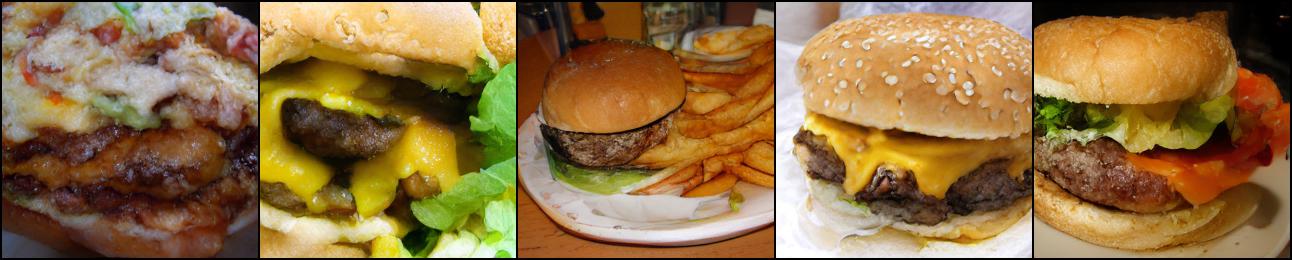}
    \includegraphics[width=\textwidth]{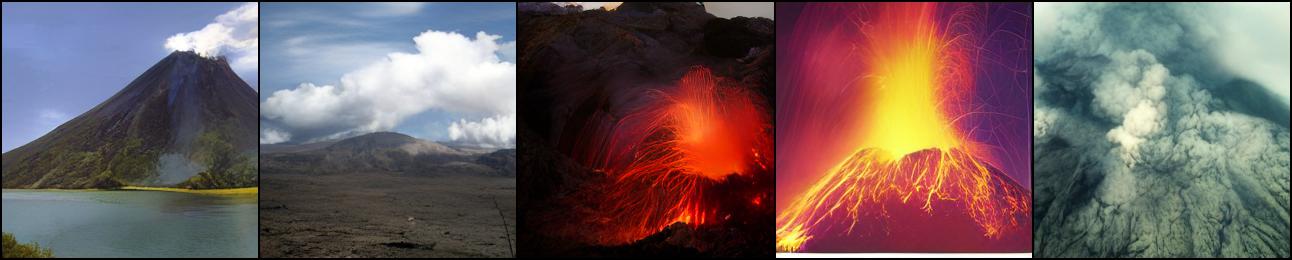}
    \caption{Additional examples of class-conditional image generation by our model with 1.4B parameters trained on ImageNet. The class conditions are Cock (7), Ostrich (9), Tibetan terrier (200), Space shuttle (812), Cheeseburger (933), and Volcano (980), respectively.}
    \label{fig:appendix_cIN}
\end{figure}

\begin{figure}
    \centering
    \begin{tabular}{m{0.18\textwidth}m{0.78\textwidth}}
         \textit{pine tree on a background of the sea$^\dagger$} & \includegraphics[width=0.78\textwidth]{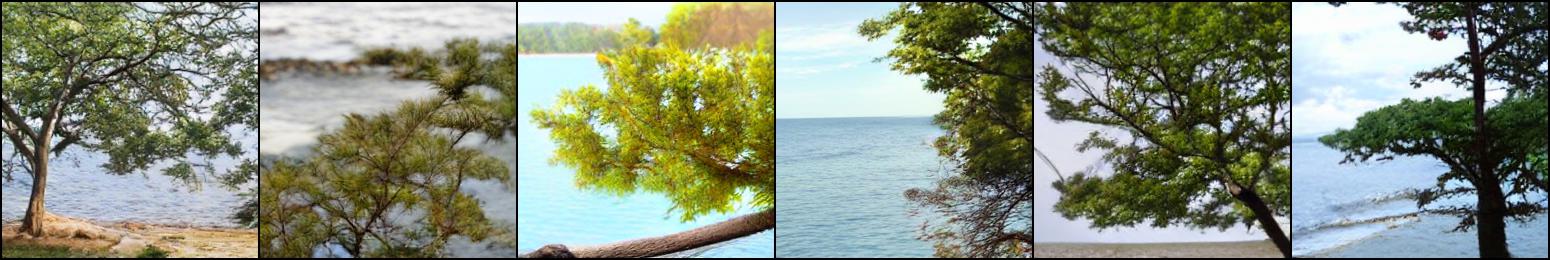} \\
         \textit{cherry blossom tree on a background of the sea} & \includegraphics[width=0.78\textwidth]{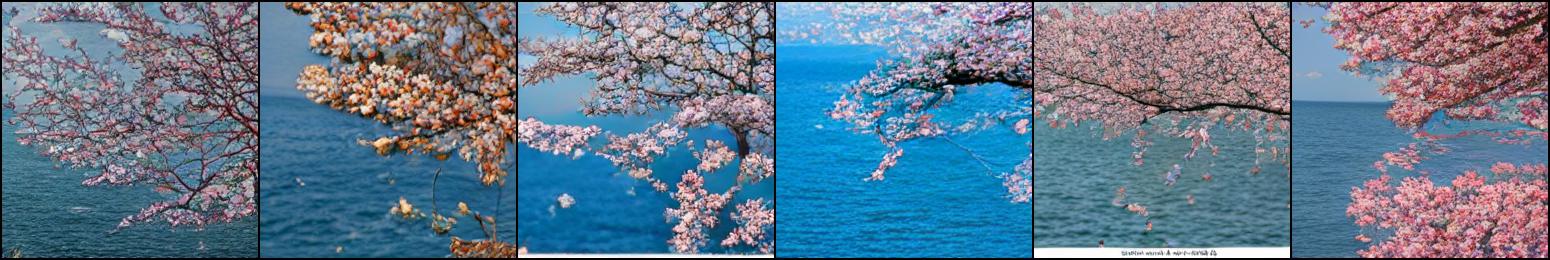} \\ \hline \\ [-0.85em]
         \textit{road on the mountain during rainy season$^\dagger$} & \includegraphics[width=0.78\textwidth]{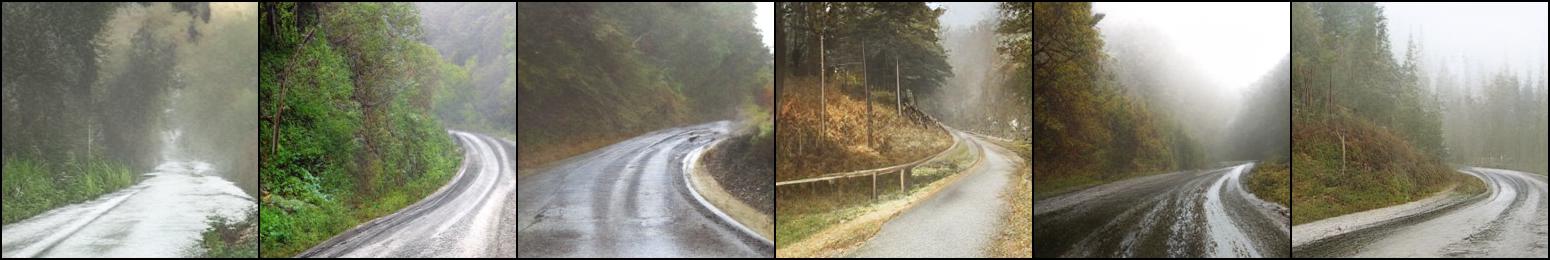} \\
         \textit{road on the mountain during snowy season} & \includegraphics[width=0.78\textwidth]{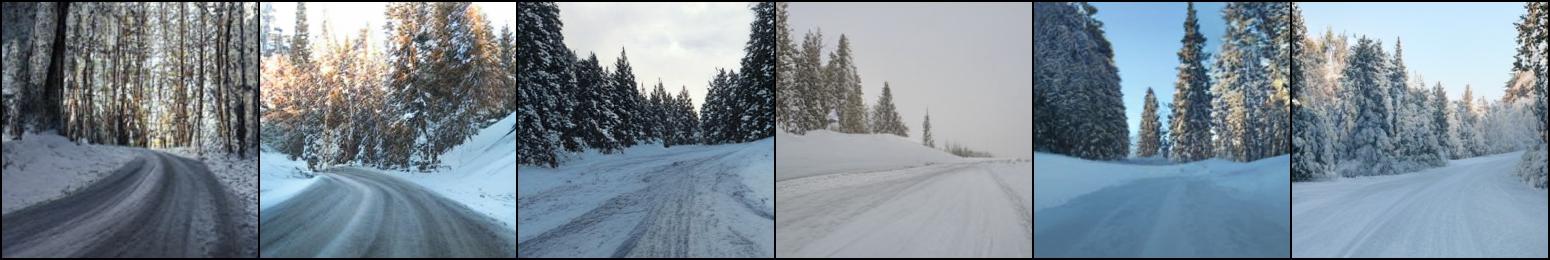} \\ \hline \\ [-0.85em]
         \textit{young brown bear in the wild forest$^\dagger$} & \includegraphics[width=0.78\textwidth]{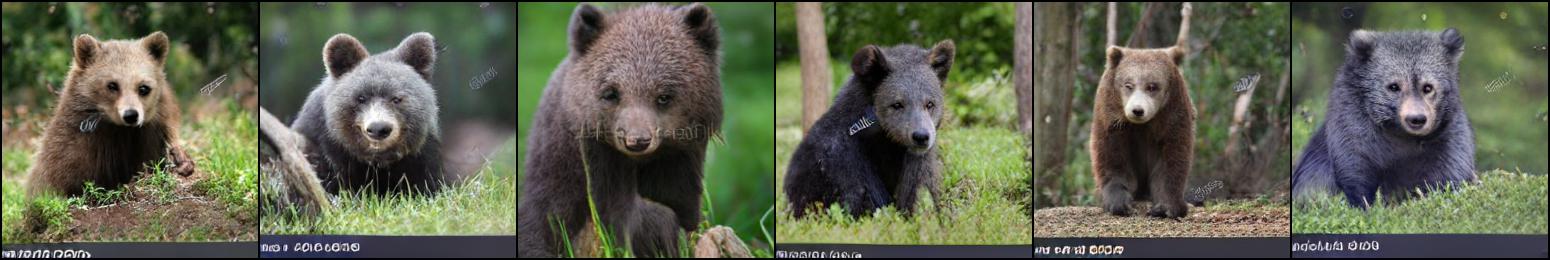} \\
         \textit{young brown bear on the seashore} & \includegraphics[width=0.78\textwidth]{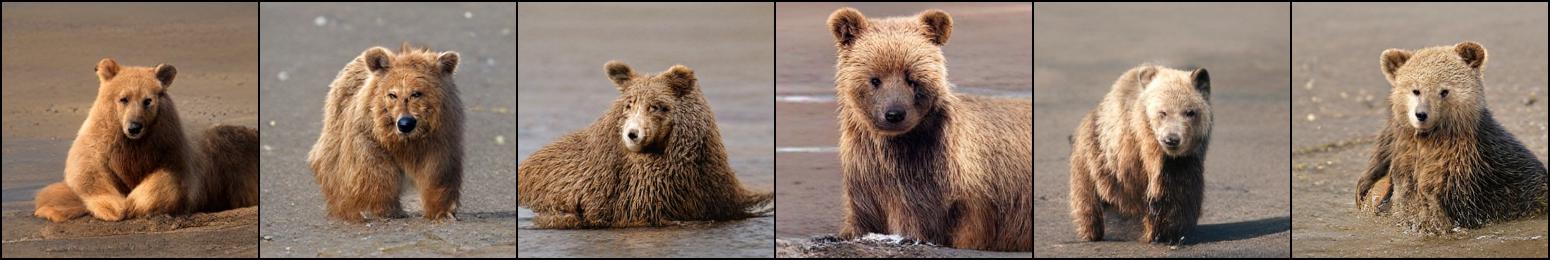} \\ \hline \\ [-0.85em]
         \textit{A cheeseburger in front of a mountain range covered with snow.} & \includegraphics[width=0.78\textwidth]{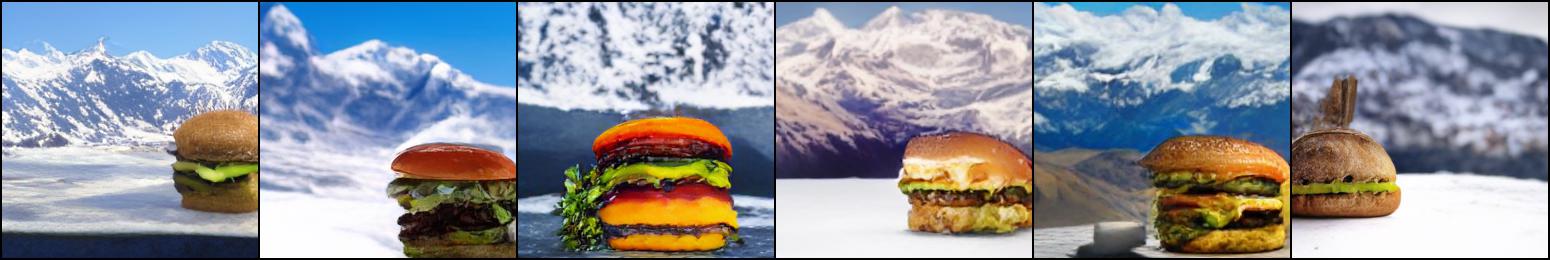} \\
         \textit{A pizza in front of a mountain range covered with snow.} & \includegraphics[width=0.78\textwidth]{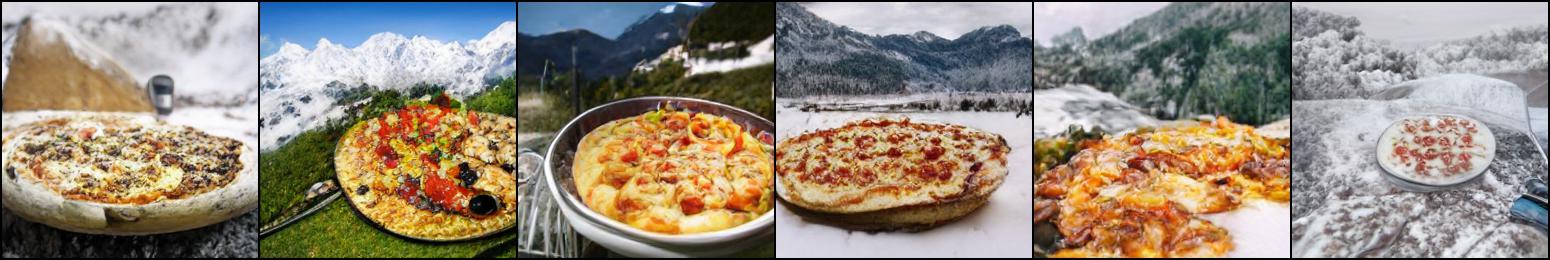} 
    \end{tabular}
    \caption{Additional examples of text-conditional image generation by our model with 654M parameters trained on CC-3M. $\dagger$ denotes that the caption exists in the validation dataset of CC-3M.}
    \label{fig:appendix_cc3m_control}
\end{figure}

\begin{figure}
    \centering
    \includegraphics[width=\textwidth]{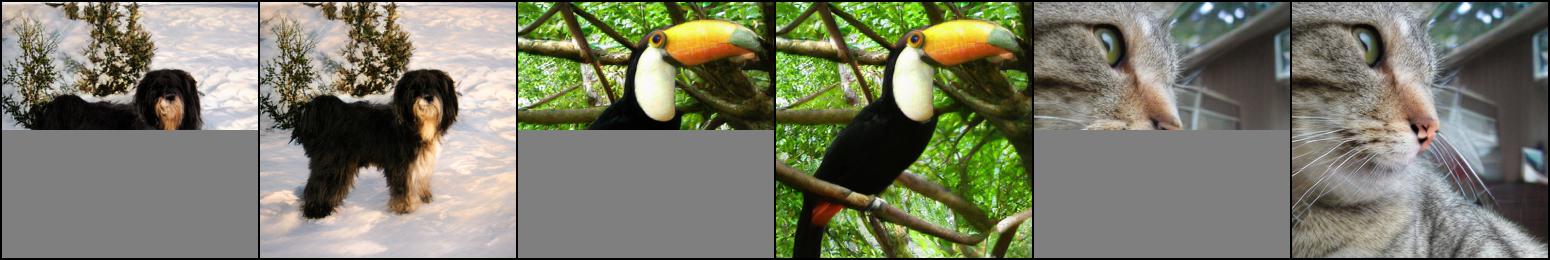}
    \includegraphics[width=\textwidth]{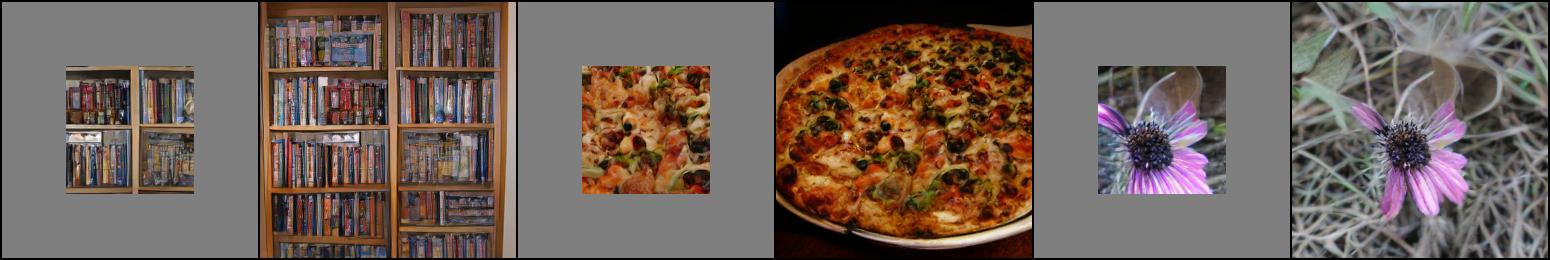}
    \includegraphics[width=\textwidth]{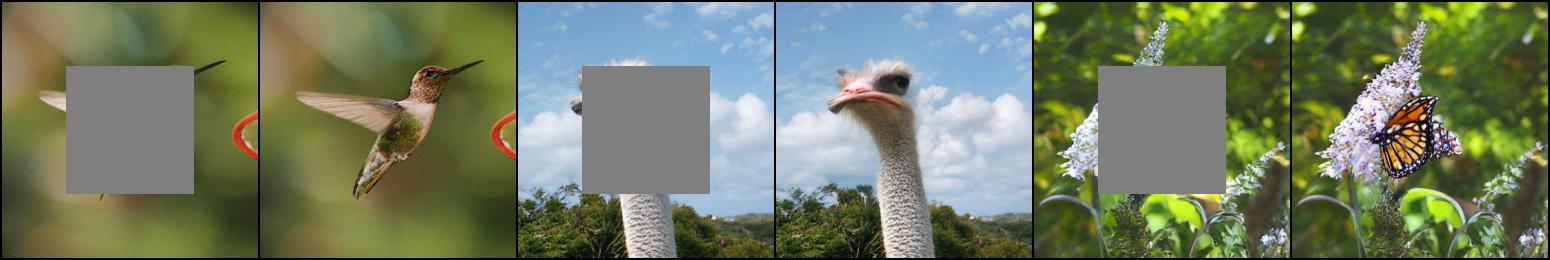}
    \includegraphics[width=\textwidth]{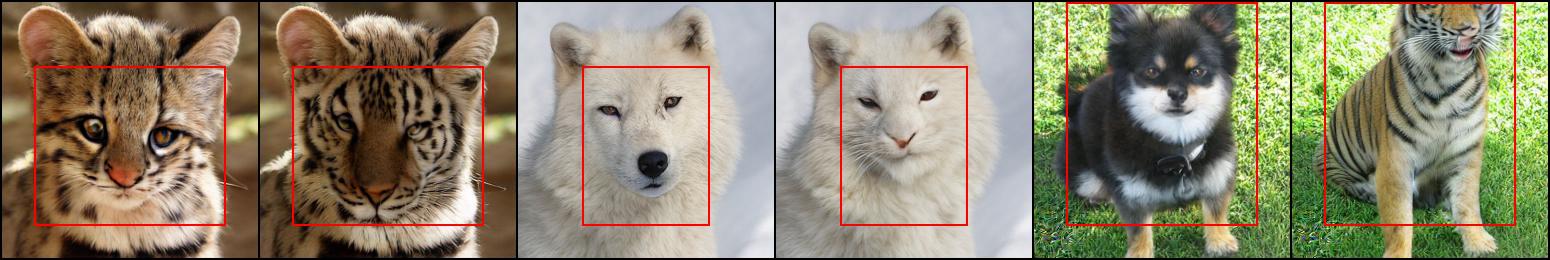}
    \caption{Additional examples of image inpainting by our model with 1.4B parameters trained on ImageNet. All masked images are taken from the validation set of ImageNet. (Top 3 rows) Inpainted images when conditioned on the class of the original image. (Bottom row) Images where the region inside the red box is inpainted with the class condition Tiger (292).}
    \label{fig:appendix_inpainting}
\end{figure}

\begin{figure}
\centering
\includegraphics[width=\textwidth]{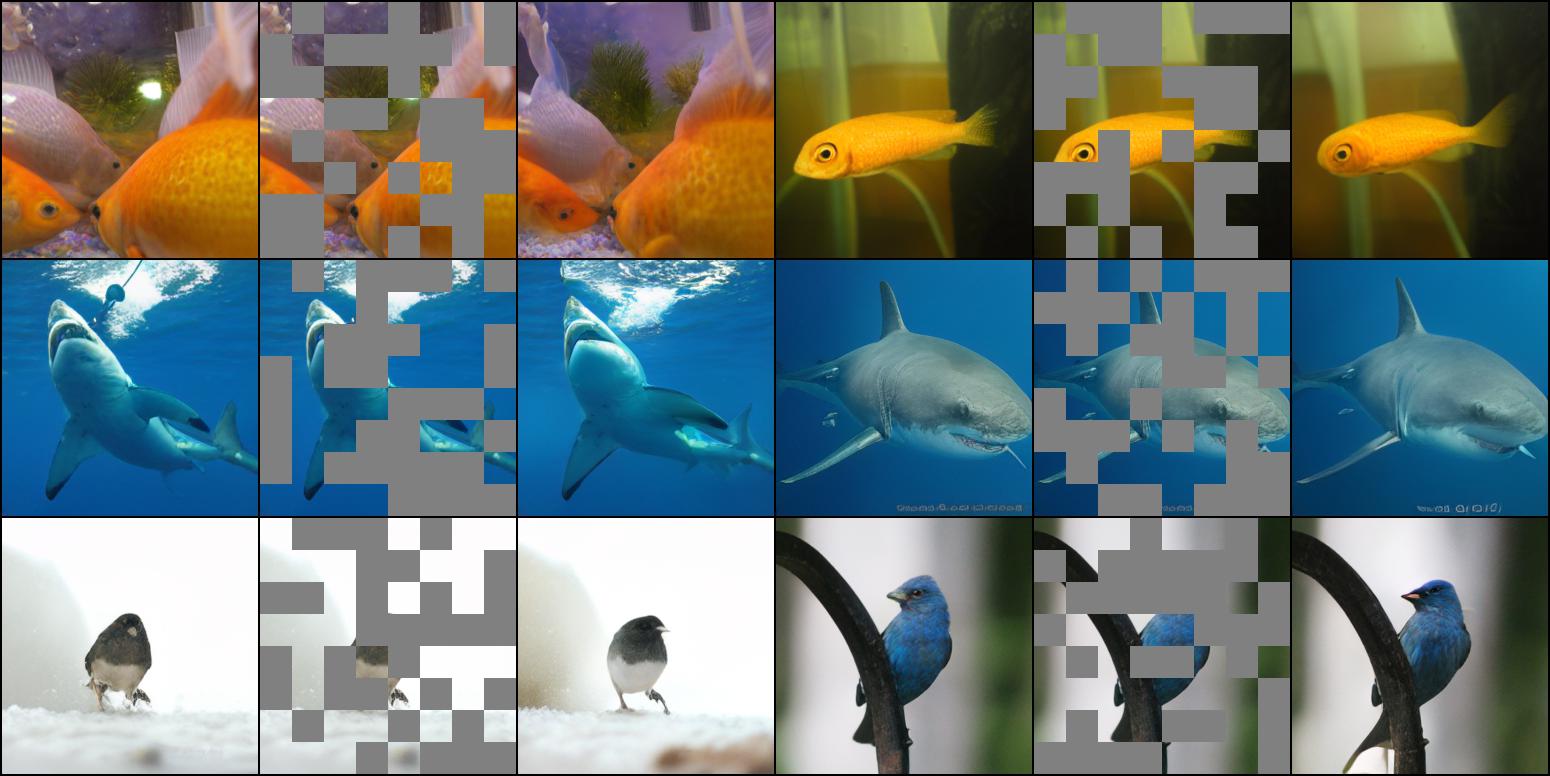}
\caption{The triplets of the original image, randomly masked image, and mask-infilled image by our model. Although half of an image is randomly masked, Contextual RQ-Transformer can infill the masks and remains the global contexts of the original image.}
\label{fig:appendix_half_mask_recon}
\end{figure}

\section{Additional Generation Examples}
In this section, we show additional examples of generated images by our Contextual RQ-Transformer.
We use 1.4B parameters of Contextual RQ-Transformer trained on ImageNet for class-conditional image generation and 654M parameters of our model trained on CC-3M for text-conditional image generation. 

\subsection{Examples of Class-conditional Image Generation}
Figure~\ref{fig:appendix_cIN} shows the generated images on class conditions in ImageNet.
Contextual RQ-Transformer generates diverse and high-quality images conditioned on given class conditions.

\subsection{Examples of Text-conditioned Image Generation}
Figure~\ref{fig:appendix_cIN} shows the generated images by our model on various text conditions.
We use the captions in the validation dataset$^\dagger$ of CC-3M and change a keyword in the caption to validate the generalization power of our model on unseen compositions of texts.
Figure~\ref{fig:appendix_cIN} shows that our Contextual RQ-Transformer can generate high-quality images on diverse compositions of text conditions, even though the text condition is unseen during training.

\subsection{Examples of Image In-painting}
Figure~\ref{fig:appendix_inpainting} shows that Draft-and-Revise decoding can also be used to inpaint the prescribed region of a given image.
For image inpainting, a random partition $\bPi$ used in each application of $\textsc{UPDATE}$ is set to be the partition of the masked region, instead of all spatial positions.
The first three rows of Figure~\ref{fig:appendix_inpainting} show the image inpainting results, where the original images are taken from the validation set of ImageNet, either bottom, center, or outside is masked, and the class of the original image is given as a condition. 
On the other hand, the last row of Figure~\ref{fig:appendix_inpainting} shows that our Contextual RQ-Transformer is also capable of image editing via inpainting, by conditioning on a class-condition that is not the class of the original image. 

\subsection{Examples of Mask-Infilling of Half Masked Images}
After we randomly mask the half of images in the ImageNet validation dataset, Contextual RQ-Transformer infills the masked regions in Figure~\ref{fig:appendix_half_mask_recon}.
The results show that Contextual RQ-Transformer can infill the masked regions, while preserving the global contexts of original images.
Note that the results are also aligned to the experimental results in previous approaches~\cite{MAE,MaskGIT}.
That is, our draft-and-revise decoding can preserve the global contexts of the draft images in the revise phase, although we use a small value of $T_\text{revise}$ such as 2 or 4.
Note that the fine-grained details can be changed after infilling masks, since our method randomly samples the codes for unmasking based on the predicted softmax distribution over codes in Eq. 9.


\end{document}